\documentclass[review]{elsarticle}
\makeatletter
\def\ps@pprintTitle{%
 \let\@oddhead\@empty
 \let\@evenhead\@empty
 \def\@oddfoot{\leftline{\footnotesize \textit{Accepted in Computer Vision and Image Understanding 2015}}}%
 \let\@evenfoot\@oddfoot}
\makeatother
\usepackage{lineno,hyperref}
\usepackage{graphicx}
\usepackage{graphics}
\usepackage{epsfig} 
\usepackage{gensymb}
\usepackage[cmex10]{amsmath}
\usepackage{amssymb}
\usepackage{epstopdf}
\usepackage{amsfonts}
\usepackage{algorithm}
\usepackage{algorithmic}

\journal{Computer Vision and Image Understanding}

\begin{document}

\begin{frontmatter}

\title{Factorization of View-Object Manifolds for Joint Object Recognition and Pose Estimation\tnoteref{mytitlenote}}
\tnotetext[mytitlenote]{This is an extension of the paper entitled "Joint Object and Pose Recognition Using Homeomorphic Manifold Analysis" \cite{ZhangAAAI13} that was presented at the 27th AAAI Conference on Artificial Intelligence (AAAI-13) held July 14–-18, 2013 in Bellevue, Washington, USA.}

\author[address1,address2]{Haopeng Zhang\corref{mycorrespondingauthor}}
\cortext[mycorrespondingauthor]{Corresponding author}
\ead{zhanghaopeng@buaa.edu.cn}

\author[address3]{Tarek El-Gaaly}
\ead{tgaaly@cs.rutgers.edu}

\author[address3]{Ahmed Elgammal}
\ead{elgammal@cs.rutgers.edu}

\author[address1,address2]{Zhiguo Jiang}
\ead{jiangzg@buaa.edu.cn}

\address[address1]{Image Processing Center, School of Astronautics, Beihang University, Beijing, 100191, China}
\address[address2]{Beijing Key Laboratory of Digital Media, Beijing, 100191, China}
\address[address3]{Department of Computer Science, Rutgers University, Piscataway, NJ 08854, USA}

\begin{abstract}
Due to large variations in shape, appearance, and viewing conditions, object recognition is a key precursory challenge in the fields of object manipulation and robotic/AI visual reasoning in general. Recognizing object categories, particular instances of objects and viewpoints/poses of objects are three critical subproblems robots must solve in order to accurately grasp/manipulate objects and reason about their environments. Multi-view images of the same object lie on intrinsic low-dimensional manifolds in descriptor spaces (e.g. visual/depth descriptor spaces). These object manifolds share the same topology despite being geometrically different. Each object manifold can be represented as a deformed version of a unified manifold. The object manifolds can thus be parameterized by its homeomorphic mapping/reconstruction from the unified manifold. In this work, we develop a novel framework to jointly solve the three challenging recognition sub-problems, by explicitly modeling the deformations of object manifolds and factorizing it in a view-invariant space for recognition. We perform extensive experiments on several challenging datasets and achieve state-of-the-art results.
\end{abstract}


\end{frontmatter}


\section{Introduction}


Visual object recognition is a challenging problem with many real-life applications. The difficulty of the problem is due to variations in shape and appearance among objects within the same category, as well as the varying viewing conditions, such as viewpoint, scale, illumination, etc.
Under this perceptual problem of visual recognition lie three subproblems that are each quite challenging: category recognition, instance recognition, and pose estimation.  Impressive work have been done in the last decade on developing computer vision systems for generic object recognition. Research has spanned a wide spectrum of recognition-related issues, however, the problem of multi-view recognition remains one of the most fundamental challenges to the progress of the computer vision.

The problems of object classification from multi-view setting (multi-view recognition) and pose recovery are coined together, and directly impacted by the way shape is represented.
Inspired by Marr's 3D object-centric doctrine~\cite{Marr82}, traditional 3D pose estimation algorithms often solved the recognition, detection, and pose estimation problems simultaneously ({\it e.g.}\cite{grimson1985recognition,lamdan1988geometric,lowe1987three,shimshoni1997finite}), through 3D object representations, or through invariants. However, such models were limited in their ability to capture large within-class variability, and were mainly focused on recognizing instances of objects. In the last two decades the field has shifted to study 2D representations based on local features and parts, which encode the geometry loosely  ({\it e.g.} pictorial structure like methods~\cite{Felzen05,pedro2010}), or does not encode the geometry at all ({\it e.g.} bag of words methods~\cite{bof_nineclasses04,bof_iccv05}.)  Encoding the geometry and the constraints imposed by objects' 3D structure are essential for pose estimation. Most research on generic object recognition bundle all viewpoints of a category into one representation; or learn view-specific classifiers from limited viewpoints,  {\it e.g.} frontal cars, side-view cars, rear cars, etc. Recently, there has been an increasing interest in object categorization in the multi-view setting, as well as recovering object pose in 3D, {\it e.g.}~\cite{Chiu07,Kushal07,Savarese07,Liebelt08,Silvio09,sun09,Payet2011,mei2011robust}. However, the representations used in these approaches are mainly category-specific representations, which do not support scaling up to a large number of categories.

The fundamental contribution of this paper is the way we address the problem. We look at the problem of multi-view recognition and pose estimation as a style and content separation problem, however, in an unconventional and unintuitive way. The intuitive way is to model the category as the content and the viewpoint as a style variability.  Instead, we model the viewpoint as the content and the category as a style variability. This unintuitive way of looking at the problem is justified from the point  of view of learning the visual manifold of the date. The manifold of different views of a given object is intrinsically low in dimensionality, with known topology. Moreover, we can show that view manifolds of all objects are deformed version of each other. In contrast, the manifold of all object categories is hard to model given all within-class variability of objects and the enormous number of categories.  Therefore, we propose to model the category as a ``style'' variable over the view manifold of objects. We show that this leads to models that can untangle the appearance and shape manifold of objects, and lead to multi-view recognition.
The formulation in this paper is based on the concept of Homeomorphic Manifold Analysis (HMA)~\cite{HMA}. Given a set of topologically equivalent manifolds, HMA models the variation in their geometries in the space of functions that maps between a topologically-equivalent common representation and each of them. HMA is based on decomposing the style parameters in the space of nonlinear functions that map between a unified embedded representation of the content manifold and style-dependent visual observations.
In this paper, we adapt a similar approach to the problem of object recognition, where we model the viewpoint as a continuous content manifold and separate object style variables as view-invariant descriptors for recognition. This results in a generative model of object appearance as a function of multiple latent variables, one describing the viewpoint and lies on a low-dimensional manifold, and the other describing the category/instance and lies on a low-dimensional subspace.
A fundamental different in our proposed framework is the way 3D shape is encoded. An object's 3D shapes imposes deformation of its view manifold. Our framework, explicitly models the deformations of object manifolds and factorizes it in a view-invariant space for recognition. It should be notice that we ignore the problem of detection/localization in this paper, and only focus on the problem of recognition and pose estimation assuming that bounding boxes or masks of the objects are given.

Pose recognition/estimation is fundamentally a six-degree-of-freedom (6DoF) problem~\cite{tamjidi2013}, including 3DoF position $[x,y,z]$ and 3DoF orientation $[yaw, pitch, roll]$. However, in practical computer vision and robotic applications, pose estimation typically means solving for the some or all of the orientation degrees of freedom, while solving for the 3DoF position is usually called \emph{localization}. In this paper, we focused on the problem of estimating the 3DoF orientation of the object (or the 3DoF viewing orientation of the camera relatively), i.e. we assumed the camera looking at the object in a fixed distance. We firstly considered the case of 1DoF orientation, i.e. a camera looking at an object on a turntable setting, which results in a one-dimensional view manifold, and then generalized to 2DoF and 3DoF orientation. Generalization to recover the full 6DoF of a camera is not obvious. Recovering the full 6DoF camera pose is possible for a given object instance, which can be achieved by traditional model-based method. However, this is a quite challenging task for the case of generic object categories. There are various reasons why we only consider 3DoF viewing orientation and not full 6DoF. First, it quite hard to have training data that covers the space of poses in that case; all the state-of-the-art dataset are limited to only a few views, or at most, multiple views of an object on a turn-table with a couple of different heights. Second, practically, we do not see objects in all possible poses, in many applications the poses are quite limited to a viewing circle or sphere. Even humans will have problems recognizing objects in unfamiliar poses. Third, for most applications, it is not required to know the 6DoF pose, 1DoF pose is usually enough. Definitely for categorization 6DoF is not needed.  In this paper we show that we can learn from a viewing circle and generalize very well to a large range of views around it.

The rest of this paper is organized as follows. Section \ref{Sec:relatework} discusses the related work, and Section \ref{Sec:framework} summarizes our factorized model and its application to joint object and pose recognition. Separately, Section \ref{Sec:learning} and Section \ref{Sec:inference} describe how to learn the model and how to use this model to infer for category, instance and pose in detail. Section \ref{Sec:exp} evaluates the model and compares it to other state-of-the-art methods. Finally, Section \ref{Sec:concl} concludes the paper.


\section{Related Work}
\label{Sec:relatework}


\subsection{Recognition and Pose Estimation}

Traditional 3D pose estimation algorithms often solve the recognition and pose estimation problems simultaneously using 3D object model-bases, hypothesis and test principles, or through the use of invariants, e.g.~\cite{grimson1985recognition,lamdan1988geometric,lowe1987three,shimshoni1997finite}. Such models are incapable of dealing with large within-class variability and have been mainly focused on recognizing instances previously seen in the model-base. This limitation led to the development, over the last decade, of very successful categorization methods mainly based on local features and parts. Such methods loosely encode the geometry, e.g. methods like pictorial structure \cite{Felzen05}; or does not encode the geometry at all, e.g. bag of words \cite{bof_nineclasses04,bof_iccv05}.



There is a growing recent interest in developing representations that captures 3D geometric constraints in a flexible way to handle the categorization problem. The work of Savarese and Fei-Fei~\cite{Savarese07,Savarese08} was pioneering in that direction. In~\cite{Savarese07,Savarese08}  a part-based model was proposed where canonical parts are learned across different views, and a graph representation is used to model the object canonical parts. Successful recent work have proposed learning category-specific detection models that is able to estimate object pose (e.g.~\cite{mei2011robust,Payet2011,schels2012learning,Pepik12}). This has an adverse side-effect of not being scalable to a large number of categories and dealing with high within-class variation. Typically papers on this area focus on evaluating the detection and pose estimation performance and do not evaluate the categorization performance. In contrast to category-specific representations, we focus on developing a common representation for recognition and pose estimation.  This achieved through learning a view-invariant representation using a proposed three-phase process that can use images and videos in a realistic learning scenario.

Almost all the work on pose estimation and multi-view recognition from local features is based on formulating the problem as a classification problem where view-based classifiers and/or viewpoint classifiers are trained. These \emph{classification}-based approaches solve pose estimation problem in a discrete way simultaneously or not with recognition problem. They use several discrete (4, 8, 16 or more) view-based/pose-based classifiers, and take the classification results as the estimated poses. For example, in~\cite{Guo2008}, 93 support vector machine (SVM) classifiers were trained. It is obvious that only discrete poses can be obtained by these classification-based methods, and the accuracy depends on the number of classifiers. On the other hand, there are also works formulate the problem of pose estimation as a regression problem by learning the regression function within a specific category, such as car or head, e.g. \cite{Ando2005,Torki11,Fanelli2011headpose,ElGaaly2012rgbd_mkl,Pan2013headpose}. These \emph{regression}-based approaches solve pose estimation in a continuous way, and can provide continuous pose prediction. A previous comparable study in~\cite{Guo2008} shows that the regression method (i.e. support vector regression, SVR) performs well in either horizontal or vertical head pose variations comparing to SVM classifiers. More recent regression-based approaches~\cite{Torki11,ElGaaly2012rgbd_mkl} also report better pose estimation results than classification-based methods on some challenging datasets. Generally, pose estimation is essentially a continuous problem, since the pose varies continuously in real world. Thus, continuously estimating the poses is more conformable to the essence of the problem.

In the domain of \\
modal data, recent work by \cite{Lai2011optree} uses synchronized multi-modal photometric and depth information (i.e. RGB-D) to achieve significant performance in object recognition. They build an object-pose tree model from RGBD images and perform hierarchical inference. Although performance of category and instance recognition is significant, object pose recognition performance is less so. The reason is the same: a classification strategy for pose recognition results in coarse pose estimates and does not fully utilize the information present in the continuous distribution of descriptor spaces. In the work by \cite{Fanelli2011headpose,Fanelli2013randomforests}, random regression forests were used for real time head pose estimation from depth images, and such a continuous pose estimation method can get 3D orientation errors less than $10^\circ$ respectively.

\subsection{Modeling Visual Manifolds for Recognition}
Learning image manifolds has been shown to be useful in recognition, for example for learning appearance manifolds from different views~\cite{murase95visual}, learning activity and pose manifolds for activity recognition and tracking~\cite{Elgammal04CVPRb,Urtasun06CVPR}, etc.
The seminal work of  Murase and Nayar~\cite{murase95visual} showed how linear dimensionality reduction using PCA~\cite{Jolliffe86PCA} can be used to establish a representation of an object's view and illumination manifolds. Using such representation, recognition of a query instance can be achieved by searching for the closest manifold. However, such a model is mainly a projection of the data to a low-dimensional space and does not provide a way to untangle the visual manifold.  The pioneering work of Tenenbaum and Freeman~\cite{Tenenbaum00} formulated the separation of style and content using  a bilinear model framework. In that work, a bilinear model was used to decompose face appearance into two factors: head pose and different people as style and content interchangeably.
They presented a computational framework for model fitting using SVD.  A bilinear model is a special case of a more general multilinear model.
In~\cite{Vasilescu02}, multilinear tensor analysis was used to decompose face images into orthogonal factors controlling the appearance of the face including geometry (people), expressions, head pose, and illumination using High Order Singular Value Decomposition (HOSVD)~\cite{Lathauwer00JMAAa}. N-mode analysis of higher-order tensors was originally proposed and developed in~\cite{Kapteyn86PSY,Magnus88} and others. A fundamental limitation with bilinear and multilinear models is that they need an aligned product space of data (all objects $\times$ all views $\times$ all illumination etc.).

The proposed framework utilizes bilinear and multilinear analysis. However, we use such type of analysis in a different way that avoids their inherent limitation. The content manifold, which is the view manifold in our case, is explicitly represented using an embedded representation, capitalizing in the knowledge of its dimensionality and topology. Given such representation, the style parameters are factorized in the space of nonlinear mapping functions between a representation of the content manifold and the observations. The main advantage of this approach is that, unlike bilinear and multilinear models that mainly discretize the content space, the content in our case is treated as a continuous domain, and therefore aligning of data is not needed.

The introduction of nonlinear dimensionality reduction techniques such as Local Linear Embedding (LLE)~\cite{Roweis00},  Isometric Feature Mapping (Isomap)~\cite{Tenenbaum98}, and others~\cite{Belkin03NC,Brand03,lawrence03NIPS,Weinberger04CVPR}, provide tools to represent complex manifolds in low-dimensional embedding spaces, in ways that aim at preserving the manifold geometry. However, in practice, away from toy examples, it is hardly the case that various orthogonal perceptual aspects can be shown to correspond to certain directions or clusters in the embedding space. In the context of generic object recognition, direct dimensionality reduction of visual features was not shown to provide an effective solution; to the contrast, the state of the art is dominated by approaches that rely on extremely high-dimensional feature spaces to achieve class linear separability, and the use of discriminative classifier, typically SVM, in these spaces. By learning the visual manifold, we are not advocating for a direct dimensionality reduction solution that mainly just project data aiming at preserving the manifold geometry locally or globally. We are arguing for a solution that is able to factorize and untangle the complex visual manifold to achieve multi-view recognition.

\section{Framework}
\label{Sec:framework}


\subsection{Intuition}
The objective of our framework is to learn a manifold representation for multi-view objects that supports category, instance and viewpoint recognition. In order to achieve this, given a set of images captured from different viewpoints, we aim to learn a generative model that explicitly factorizes the following:
\begin{itemize}
\item Viewpoint variable (within-manifold parameterization): smooth parameterization of the viewpoint variations, invariant to the object's category.
\item Object variable (across-manifold parameterization): parameterization at the level of each manifold that characterizes the object's instance/category, invariant to the viewpoint.
\end{itemize}

\begin{figure}[t]
  \centering
  \includegraphics[width=4in, height=2in]{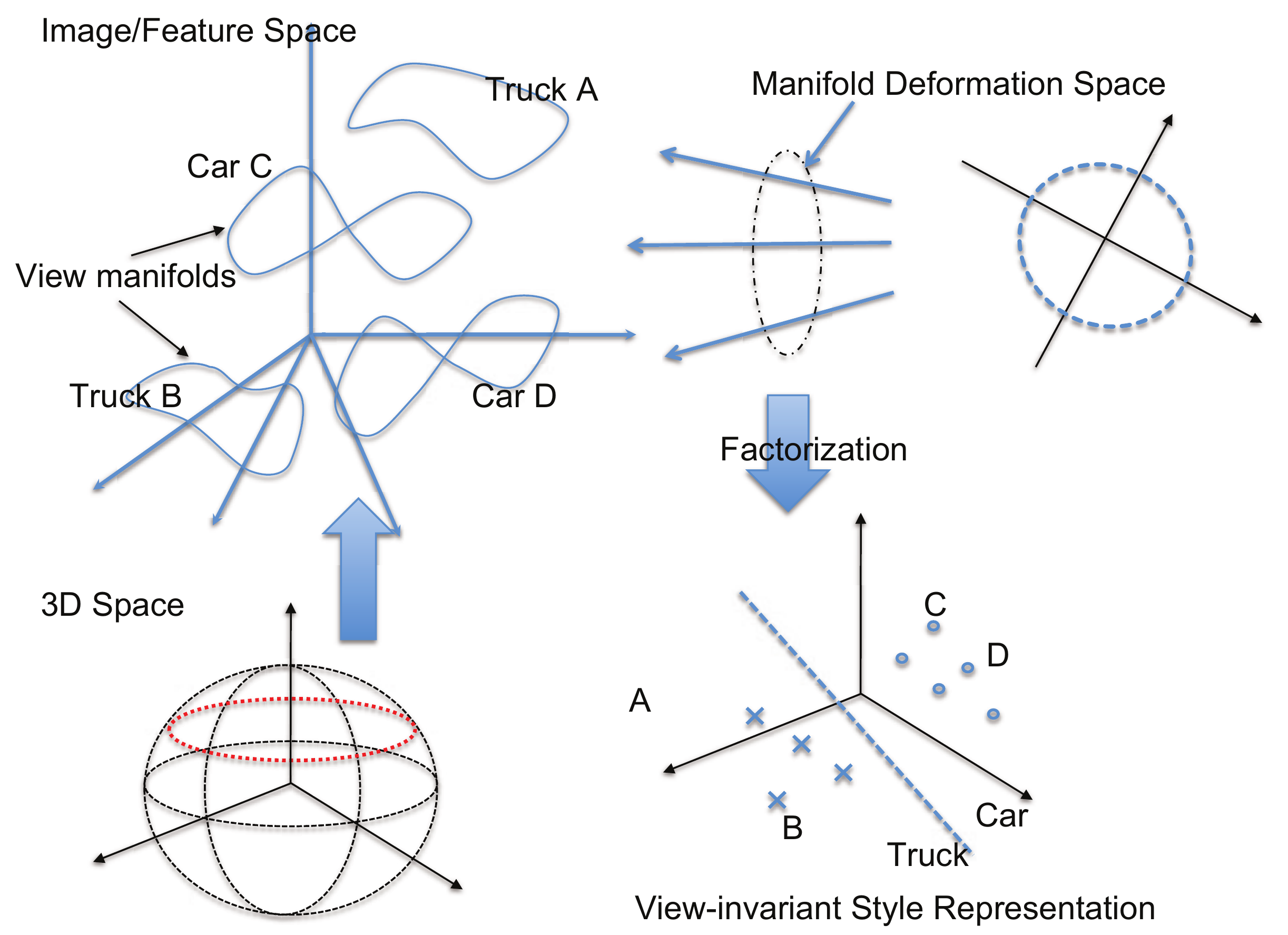}
\caption { Framework for factorizing the view-object manifold. }
  \label{F:Framework}
\end{figure}

Consider collections of images containing instances of different object classes and  different views of each instance.   The shape and appearance of an object in a given image is a function of its category, style within category, viewpoint, besides other factors that might be nuisances for recognition. Our discussion do not assume any specific feature representation of the input, we just assume that the images are vectors in some {\em input space}. The visual manifold given all these variability collectively is impossible to model. Let us first simplify the problem. Let us assume that the object is detected in the training images (so there is no 2D translation or in-plane rotation manifold). Let us also assume we are dealing with rigid objects (to be relaxed), and ignore the illumination variations (assume using an illumination invariant feature representation). Basically, we are left with variations due to category, within category, and viewpoint, i.e., we are dealing with a combined {\em view-object manifold}.

The underlying principle is that multiple views of an object lie on an intrinsic low-dimensional manifolds  in the input space (denoted as {\em view manifold}). The view manifolds of different objects are distributed in input space. To recover the category, instance and pose of a test image we need to know which manifold this image belongs to and the intrinsic coordinates of that image within the manifold. This basic view of object recognition and pose estimation is not new, and was used in the seminal work of \cite{murase95visual}. PCA~\cite{Jolliffe86PCA} was used to achieve linear dimensionality reduction of the visual data, and the manifolds of different object were represented as parameterized curves in the embedding space. However, dimensionality reduction techniques, whether linear or nonlinear, will just project the data preserving the manifold local or global geometry, and will not be able to achieve the desired untangled representation.

What is novel in our framework, is that we use the view manifold deformation as an invariant that can be used for categorization and modeling the within-class variations. Let us consider the case where different views are obtained from a viewing circle, e.g. camera viewing an object on a turntable. The view manifold of the object is a 1D closed manifold embedded in the input space.  That simple closed curve deforms in the input space depending on the object shape and appearance. The view manifold can be degenerate, e.g. imaging a textureless sphere from different views results in the same image, i.e. the visual manifold in this case is degenerate to a single point. Therefore, capturing and parameterizing the deformation of a given object's view manifold tells us information about the object category and within category variation. If the views are obtained from a full or part of the view-sphere centered around the object, it is clear that the resulting visual manifold should be a deformed sphere as well (assuming the cameras are facing toward the object).

Let us denote the view manifold of an object instance $s$ in the input space by $\mathcal{D}^s \subset \mathbb{R}^D$. $D$ is the dimensionality of the input space. Assuming that all manifolds $\mathcal{D}^s$ are not degenerate (we will discuss this issue shortly), then they are all topologically equivalent,  and homeomorphic to each other\footnote{A function $f:X \rightarrow Y$ between 2 topological spaces is called a homeomorphism if it is a bijection, continuous, and its inverse is continuous. In our case the existence of the inverse is assumed but not required for computation, i.e., we do not need the inverse for recovering pose. We mainly care about the mapping in a generative manner from $\mathcal{M}$ to $\mathcal{D}^s$. }. Moreover, suppose we can achieve a common view manifold representation across all objects, denoted by $\mathcal{M} \subset \mathbb{R}^e$, in an Euclidean embedding space of dimensionality $e$. All manifolds $\mathcal{D}^s$ are also homeomorphic to $\mathcal{M}$. In fact all these manifolds are homeomorphic to a unit circle in 2D for the case of a viewing circle, and a unit sphere in 3D for the case of full view sphere.
{\em In general, the dimensionality of the view manifold of an object is bounded by the dimensionality of viewing manifold (degrees of freedom imposed by the camera-object relative pose). }

\subsection{Manifold Parameterization}
We can achieve a parameterization of each manifold deformation by learning object-dependent regularized mapping functions $\gamma_s(\cdot):\mathbb{R}^e \to \mathbb{R}^D$  that map from $\mathcal{M}$ to each $\mathcal{D}^s$.
Given a Reproducing Kernel Hilbert Space (RKHS) of functions and its corresponding kernel $K(\cdot,\cdot)$, from the representer theorem~\cite{Kimeldorf1970:representerThrm,Poggio1990:GRBF} it follows that such functions admit a representation of the form
\begin{equation}
\label{eq:learned_style_model}
 \gamma_s (\boldsymbol{x}) = \boldsymbol{C}^s \cdot \psi (\boldsymbol{x}) \;,
\end{equation}
where $\boldsymbol{C}^s$ is a $D \times  N_{\psi}$ mapping coefficient matrix, and $\psi(\cdot):\mathbb{R}^e \to \mathbb{R}^{N_{\psi}}$ is a nonlinear kernel map, as will be described in Section~\ref{Sec:learning}.

In the mapping (Eq.~\ref{eq:learned_style_model}), the geometric deformation of manifold $\mathcal{D}^s$, from the common manifold $\mathcal{M}$, is encoded in the coefficient matrix $\boldsymbol{C}^s$. Therefore, the space of matrices $\{ \boldsymbol{C}^s \}$ encodes the variability between different object manifolds, and can be used to parameterize such manifolds.  We can parameterize the variability across different manifolds in a subspace in the  space of coefficient matrices. This results in a generative model in the form
\begin{equation}
\label{eq:model}
\gamma (\mathbf{x},\mathbf{s}) = \mathcal{A} \times_2 \mathbf{s} \times_3 \psi (\mathbf{x}).
\end{equation}
In this model $\mathbf{s} \in \mathbb{R}^{d_s} $ is a parameterization of manifold $\mathcal{D}^s$ that signifies the variation in category/instance of an object. $\mathbf{x}$ is a representation of the viewpoint that evolves around the common manifold $\mathcal{M}$. $ \mathcal{A}$ is a third order tensor of dimensionality $D \times d_s \times N_{\psi}$, where $\times_i$ is the mode-i tensor product as defined in~\cite{Lathauwer00JMAAa}. In this model, both the viewpoint and object latent representations, $\mathbf{x}$ and $\mathbf{s}$,  are continuous.

There are several reasons why we learn the mapping in a generative manner from  $\mathcal{M}$ to each object manifold (not the other way). First, this direction guarantees that the mapping is a function, even in the case of degenerate manifolds (or self intersections) in the input space. Second, mapping from a unified representation as  $\mathcal{M}$ results in a common RKHS of functions. All the mappings will be linear combinations of the same finite set of basis functions. This facilitates factorizing the manifold geometry variations in the space of coefficients in Eq ~\ref{eq:model}.

Given a test image $\mathbf{y}$ recovering the category, instance and pose reduces to an inference problem where the goal is to find $\mathbf{s}^*$ and $\mathbf{x}^*$ that minimizes a reconstruction error, i.e.,
\begin{equation}
\arg \min_{\mathbf{s},\mathbf{x}}\| \mathbf{y}- \mathcal{A} \times_2 \mathbf{s} \times_3 \psi (\mathbf{x})\|^2.
\label{eq:rec_err}
\end{equation}
Once $\mathbf{s}$ is recovered, an instance classifier and a category classifier can be used to classify $\mathbf{y}$.

Learning the model is explained in Section~\ref{Sec:learning}. Here we discuss and justify our choice of the common manifold embedded representation. Since we are dealing with 1D closed view manifolds, an intuitive common representation for these manifolds is a unit circle in $\mathbb{R}^2$. A unit circle has the same topology as all object view manifolds (assuming no degenerate manifolds), and hence, we can establish a homeomorphism between it and each manifold.

Dimensionality reductions (DR) approaches, whether linear (such as PCA \cite{Jolliffe86PCA} and PPCA~\cite{Tipping99PPCA})  or nonlinear (such as  isometric feature mapping (Isomap)~\cite{Tenenbaum98}, Locally linear embedding (LLE) \cite{Roweis00}, Gaussian Process Latent Variable Models (GPLVM)~\cite{lawrence03NIPS})  have been widely used for embedding manifolds in low-dimensional Euclidean spaces. DR approaches find an optimal embedding (latent space representation) of a manifold by minimizing an objective function that preserves local (or global) manifold geometry. Such low-dimensional latent space is typically used for inferring object pose or body configuration. However, since each object has its own view manifold, it is expected that the embedding will be different for each object. On the other hand, using DR to embed data from multiple manifolds together will result in an embedding dominated by the inter-manifold distance and the resulting representation cannot be used as a common representation.

Embedding multiple manifolds using DR can be achieved using manifold alignment, e.g. \cite{Ham05}. If we embed aligned view manifolds for multiple objects where the views are captured from a viewing circle, we observe that the resulting embedding will converge to a circle. Similar results were shown in (\cite{Torki11}), where a view manifold is learned from local features from multiple instances with no prior alignment. This is expected since each object view manifold is a 1D closed curve in the input space, i.e. a deformed circle. Such deformation depends on object geometry and appearance. Hence it is expected that the latent representation of multiple aligned manifolds will converge to a circle. This observation empirically justifies the use of a unit circle as a general model of object view manifold in our case. Unlike DR where the goal is to find an optimal embedding that preserves the manifold geometry, in our case we only need to preserve the topology while the geometry is represented in the mapping space. This facilitates parameterizing the space of manifolds. Therefore, the unit circle represents an \emph{ideal} conceptual manifold representation, where each object manifold is a deformation of that ideal case. In some sense we can think of a unit circle as a prior model for all 1D view manifolds. If another degree of freedom is introduced which, for example, varies the pitch angle of the object on the turn-table then a sphere manifold would capture the conceptual geometry of the pose and be topologically-equivalent.

\paragraph{Dealing with degeneracy} Of course the visual manifold can be degenerate in some cases or it can be self intersecting,  because of the projection from 3D to 2D and lack of visual features, e.g., images of a textureless sphere. In such cases the homeomorphic assumption does not hold. The key to tackle this challenge is in learning the mapping in a generative manner from $\mathcal{M}$ to $\mathcal{D}^s$, not in the other direction. By enforcing the known non-degenerate topology on  $\mathcal{M}$, the mapping from $\mathcal{M}$ to $\mathcal{D}^s$ still exists, still is a function, and still captures the manifold deformation. In such cases the recovery of object pose might be ambiguous and ill-posed. In fact, such degenerate cases can be detected by rank-analysis of the mapping matrix $\boldsymbol{C}^s$.

\section{Learning the Model}
\label{Sec:learning}


The input to the learning algorithm is images of different objects from different viewpoint, with viewpoint labels, and category label. For learning the representation, only the viewpoint labels are needed, while the category labels are used for learning classifiers on top of the learned representation, i.e. learning the representation is ``unsupervised'' from category perspective. Images of the same object from different views is dealt with as a set of points sampled from its view manifold. The number of sampled views do not necessarily be the same, nor they have to be aligned. We first describe constructing a common ``conceptual'' view manifold representation $\mathcal{M}$ then we describe learning the model.

\subsection{View manifold representation}
\label{SSec:viewmani}
Let the sets of input images be $Y^k = \{(\mathbf{y}^k_i \in \mathbb{R}^D, \mathbf{p}^k_i), i=1,\cdots,N_k\}$ where $D$ is the dimensionality of the input space (i.e. descriptor space) and $p$ denotes the pose label. We construct a conceptual unified embedding space be in  $\mathbb{R}^e$,  where  $e$ is the dimensionality of the conceptual embedding space. Each input image will have a corresponding embedding coordinate defined by construction using the pose labels. We denote the embedding coordinates by  $X^k = \{\mathbf{x}^k_i \in \mathbb{R}^e, i=1,\cdots,N_k\}$.

If we assume the input is captured from a viewing circle with yaw angles (viewpoints): $\Theta = \{\theta^k_i \in [0,2\pi), i=1,\cdots,N_k\}$, then  the $k$-th image set is embedded on a unit circle such that $\mathbf{x}^k_i = \ [\cos{\theta^k_i}, \sin{\theta^k_i}] \in \mathbb{R}^2, i=1,\cdots,N_k$. By such embedding, multi-view images with 1D pose variation are represented on a conceptual manifold (unit circle in 2D), i.e. a normalized 1-sphere. For the case of a full view sphere (2D pose variation represented by yaw and pitch angles), images are represented on a unit-sphere in 3D, i.e. a normalized 2-sphere. And for the case of 3D pose variation represented by yaw, pitch and roll angles, the conceptual manifold will be a normalized 3-sphere in 4D. Generally, assuming the pose angles of the input are $\mathbf{p}^k_i = \{ (\theta^k_i, \beta^k_i, \zeta^k_i), i=1,\cdots,N_k\}$ where $\theta$, $\beta$ and $\zeta$ indicate yaw angle, pitch angle and roll angle respectively, then the embedded coordinate of the $i$-th image $\mathbf{y}^k_i$ is defined as
\begin{equation}
\label{eq:mani}
\mathbf{x}^k_i = \left\{
\begin{array}{l}
\left[\cos{\theta^k_i}, \sin{\theta^k_i}\right]^T \in \mathbb{R}^2~\text{(1D case)}\\
\left[
\begin{array}{c}
\cos{\theta^k_i} \cos{\beta^k_i}\\
\sin{\theta^k_i} \cos{\beta^k_i} \\
\sin{\beta^k_i}
\end{array}
\right]\in \mathbb{R}^3~\text{(2D case)}\\
\left[
\begin{array}{c}
\cos{\theta^k_i} \cos{\beta^k_i} \cos{\zeta^k_i}\\
\sin{\theta^k_i} \cos{\beta^k_i} \cos{\zeta^k_i}\\
\sin{\beta^k_i} \cos{\zeta^k_i}\\
\sin{\zeta^k_i}
\end{array}
\right] \in \mathbb{R}^4~\text{(3D case)}\\
\end{array}
\right.
\end{equation}
Notice that by embedding on a conceptual manifold, we just preserve the topology of the manifold, not the metric input space. For clarity and without loss of generality, we only consider 1D case when describing the learning and inferring procedures in the following parts of this section and the next.

\subsection{Homeomorphic Manifold Mapping}
\label{SSec:mapping}
Given an input set $Y^k$ and its embedding coordinates $X^k$ on a unit circle, we learn a regularized nonlinear mapping function from the embedding to the input space, i.e. a function $\gamma_k(\cdot) : \mathbb{R}^e \rightarrow \mathbb{R}^D$ that maps from embedding space, with dimensionality $e$, into the input space with dimensionality $D$.

To learn such mappings, we learn individual functions   $\mathcal{\gamma}_k^l : \mathbb{R}^e \rightarrow \mathbb{R}$ for the $l$-th dimension in the feature space.
Each of these functions minimizes a regularized loss functional in the form
\begin{equation}
\label{eq:gamma}
    \sum_i^{n^k} \left\|\mathbf{y}^k_{il} - \gamma_k^l(\mathbf{x}^k_i)\right\|^2 + \lambda \; \Omega[\gamma_k^l],
\end{equation}
where $\left\|\cdot\right\| $ is the Euclidean norm, $\Omega$ is a regularization function that enforces the smoothness in the learned function, and $\lambda$ is the regularizer that balances between fitting the training data and smoothing the learned function. From the representer theorem~\cite{Kimeldorf1970:representerThrm} we know that a nonlinear mapping function that minimizes a regularized risk criteria admits a representation in the form of linear combination of basis functions around arbitrary points $\mathbf{z}_j \in \mathbb{R}^e, j=1,\cdots, M$ on the manifold (unit circle). In particular we use a semi-parametric form for the function $\gamma(\cdot)$. Therefore, for the $l$-th dimension of the input, the function $\gamma_k^l$ is an RBF interpolant from $\mathbb{R}^e$ to $\mathbb{R}$. This takes the form
\begin{equation}
\label{poly}
\gamma_k^l(\mathbf{x}) = p^l(\mathbf{x}) + \sum^M_{j=1}  \omega^l_j \cdot \phi(|\mathbf{x}-\mathbf{z}_j|) ,
\end{equation}
where $\phi(\cdot)$ is a real-valued basis function, $\omega_j$  are real coefficients and $|\cdot|$  is the $2^{nd}$ norm in the embedding space.
$p^l$ is a linear polynomial with coefficients $c^l$, i.e. $p^l(\mathbf{x})= [1 \;\;\; \boldsymbol{\mathbf{x}}^{\scriptscriptstyle}] \cdot c^{l} $.
The polynomial part is needed for positive semi-definite kernels to span the null space in the corresponding RKHS. The polynomial part is essential for regularization with the choice of specific basis functions such as Thin-plate spline kernel~\cite{Wahba71splines}.
The choice of the centers is arbitrary (not necessarily data points). Therefore, this is a form of Generalized Radial Basis Function (GRBF) \cite{Poggio1990:GRBF}. Typical choices for the basis function include thin-plate spline, multiquadric, Gaussian\footnote{A Gaussian kernel does not need a polynomial part.}, biharmonic and tri-harmonic splines. The whole mapping can be written in a matrix form
\begin{equation}
\label{eq:wholemap}
\gamma^k(\mathbf{x}) = \mathbf{C}^k  \cdot \psi(\mathbf{x}),
\end{equation}
where $\mathbf{C}^k$ is a $D \times (M+e+1)$  dimensional matrix with the $l$-th row $[\omega^l_1, \cdots, \omega^l_M, {c^l}^T]$. The vector $\psi(x)=[\phi(|x-z_1|) \cdots \phi(|x-z_M|),1,x^T]^T$  represents a nonlinear kernel map from the embedded conceptual representation to a kernel induced space.
To ensure orthogonality and to make the problem well posed, the following condition constraints are imposed:  $\Sigma^M_{i=1} \omega_i p_j(x_i) = 0, j=1,\cdots,m$, where $p_j$  are the linear basis of $p$. Therefore, the solution for $\mathbf{C}^k$ can be obtained by directly solving the linear system:
\begin{equation}
\begin{pmatrix} \mathbf{A} & \mathbf{P}_x \\
 \mathbf{P}_t^T & \mathbf{0}_{(e+1) \times (e+1) } \end{pmatrix}_k  {\mathbf{C}^k}^T= \begin{pmatrix} \mathbf{Y}_k \\ \mathbf{0}_{(e+1) \times d} \end{pmatrix},
\end{equation}
$\mathbf{A}$, $\mathbf{P}_x$ and $\mathbf{P}_t$ are defined for the $k-th$ set of object images as: $\mathbf{A}$ is a $N_k \times M$ matrix with $\mathbf{A}_{ij}=\phi(|x^k_i - z_j|), i=1,\cdots,N_k, j=1,\cdots,M, \mathbf{P}_x$ is a $N_k \times (e+1)$ matrix with $i$-th row $[1,\mathbf{x}^{k^T}_i]$, $\mathbf{P}_t$ is $M \times (e+1)$  matrix with $i$-th row $[1,\mathbf{z}^T_i]$. $Y_k$ is a $N_k \times D$ matrix containing the input images for set of images $k$, i.e. $\mathbf{Y}_k=[\mathbf{y}_1^k,\cdots, \mathbf{y}^k_{N_k}]$. Solution for $\mathbf{C}^k$ is guaranteed under certain conditions on the basic functions used.

\subsection{Decomposition}
Each coefficient matrix $\mathbf{C}^k$ captures the deformation of the view manifold for object instance $k$. Given learned coefficients matrices $\mathbf{C}^1, \cdots,\mathbf{C}^K$ for each object instance, the category parameters can be factorized by finding a low-dimensional subspace that approximates the space of coefficient matrices. We call the category parameters/factors \emph{style} factors as they represent the parametric description of each object view manifold.

Let the coefficients be arranged as a $D \times K \times (M + e+1)$ tensor $\mathcal{C}$. The form of the decomposition we are looking for is:
\begin{equation}
\mathcal{C} = \mathcal{A} \times_2 \mathbf{S},
\end{equation}
where $A$ is a $D \times d_s \times (M+e+1)$ tensor containing category bases for the RBF coefficient space and $\mathbf{S}=[\mathbf{s}^1,\cdots,\mathbf{s}^K]$ is $d_s \times K$. The columns of $\mathbf{S}$ contain the instance/category parameterization. This decomposition can be achieved by arranging the mapping coefficients as a $ (D (M+e+1)) \times K$ matrix:
\begin{equation}
\mathbf{C} = \begin{pmatrix}
\mathbf{c}^1_1 & \cdots & \mathbf{c}^K_1  \\
 \vdots & \ddots & \vdots \\
\mathbf{c}^1_{M+e+1} & \cdots & \mathbf{c}^K_{M+e+1},
\end{pmatrix}
\label{eq_bigC}
\end{equation}
$[\mathbf{c}^k_1,\cdots,\mathbf{c}^k_{M+e+1}]$  are the columns of $\mathbf{C}^k$. Given $\mathbf{C}$, category vectors and content bases can be obtained by SVD as $\mathbf{C}=\mathbf{U}\mathbf{\Sigma}\mathbf{V}^T$. The bases are the columns of $\mathbf{U}\mathbf{\Sigma}$ and the object instance/category vectors are the rows of $\mathbf{V}$. Usually, $ (D (M+e+1)) \gg K$, so the dimensionality of instance/category vectors obtained by SVD will be $K$, i.e. $d_s=K$. The time complexity of SVD is $O(K^3)$ so here our approach scales cubically with the number of objects, and the space complexity is not much of a problem as SVD can be done on a large enough matrix containing tens of thousands of rows.

\section{Inference of Category, Instance and Pose}
\label{Sec:inference}


Given a test image $\mathbf{y} \in \mathbb{R}^D $ represented in a descriptor space, we need to solve for both the viewpoint parameterization $\mathbf{x}^*$ and the object instance parameterization $\mathbf{s}^*$ that minimize Eq.~\ref{eq:rec_err}. This is an inference problem and various inference algorithms can be used. Notice that, if the instance parameters $\mathbf{s}$ is known, Eq.~\ref{eq:rec_err} reduces to a nonlinear 1D search for viewpoint $\mathbf{x}$ on the unit circle that minimizes the error. This can be regarded as a solution for viewpoint estimation, if the object is known. On the other hand, if $\mathbf{x}$ is known, we can obtain a least-square closed-form approximate solution for $\mathbf{s}^*$. An EM-like iterative procedure was proposed in \cite{elgammal04cvpr} for alternating between the two factors. If dense multiple views along a view circle of an object are available, we can solve for $\mathbf{C}^*$ in Eq.~\ref{eq:wholemap} and then obtain a closed-form least-square solution for the instance parameter $\mathbf{s}^*$ as
\begin{equation}
\mathbf{s}^* = \arg \min_{\mathbf{s}}  || \mathbf{C}^* - \mathcal{A} \times_2 \mathbf{s} ||.
\end{equation}

In the case where we need to solve for both $\mathbf{x}$ and $\mathbf{s}$, given a test image, we use a sampling methods similar to particle filters \cite{Arulampalam02pf} to solve the inference problem (with $K$ category/style samples $\mathbf{s}^1,\mathbf{s}^2,\cdots,\mathbf{s}^K$ in the category/style factor space and $L$ viewpoint samples $\mathbf{x}^1,\mathbf{x}^2,\cdots,\mathbf{x}^L$ on the unit circle). We use the terms \emph{particle} and \emph{sample} interchangeably in our description of the approach. 

To evaluate the performance of each particle we define the likelihood of a particle $(\mathbf{s}^k, \mathbf{x}^l)$ as
\begin{equation}
\label{eq:w_lik}
w_{kl} = exp{\frac{-||\mathbf{y}- \mathcal{A} \times_2 \mathbf{s}^k \times_3 \psi(\mathbf{x}^l)||^2}{ 2 \sigma^2}}.
\end{equation}
It should be noticed that such a likelihood depends on the reconstruction error to be minimized in Eq.~\ref{eq:rec_err}. The less the reconstruction error is, the larger the likelihood will be. 

We marginalize the likelihood to obtain the weights for $\mathbf{s}^k$ and $\mathbf{x}^l$ as
\begin{equation}
\label{eq:weight}
W_{s^k} = \frac{\sum^L_{l=1} w_{kl} }{ \sum^K_{k=1} \sum^L_{l=1} w_{kl}}, W_{x^l} = \frac{\sum^K_{k=1} w_{kl}}{ \sum^K_{k=1} \sum^L_{l=1} w_{kl}}.
\end{equation}
Style samples are initialized as the $K$ style vectors learned by our model (decomposed via SVD of matrix $\mathbf{C}$ in Eq. \ref{eq_bigC}), and the L viewpoint samples are randomly selected on the unit circle. 

In order to reduce the reconstruction error, we resample style and viewpoint particles according to $W_s$ and $W_x$ from Normal distributions, i.e. more samples are generated around samples with high weights in the previous iteration. To keep the reconstruction error decreasing, we keep the particle with the minimum error at each iteration. Algorithm~\ref{alg:pf} summarizes our sampling approach.

In the case of classification and instance recognition, once the parameters $\mathbf{s}^*$ are known, typical classifiers, such as $k$-nearest neighbor classifier, SVM classifier, etc., can be used to find the category or instance labels. Given $\mathbf{x}^*$ on the unit circle, the exact pose angles can be computed by the inverse trigonometric function as
\begin{equation}
\theta^*=\arctan (x^*_2/x^*_1),
\end{equation}
where $x^*_1$ and $x^*_2$ are the first and second dimensions of $\mathbf{x}^*$ respectively. Similar solutions can be solved for 2D or 3D case.

\begin{algorithm}[htb]
\renewcommand{\algorithmicrequire}{\textbf{Input:}}
\renewcommand{\algorithmicensure}{\textbf{Output:}}
\caption{ Sampling approach for style and viewpoint inference.}
\label{alg:pf}
\begin{algorithmic}[1]
\REQUIRE ~~\\
    Testing image or image feature, $\mathbf{y}$;\\
    Core tensor in Eq.~\ref{eq:model}, $\mathcal{A}$;\\
    Iteration number, $IterNo$;
\ENSURE ~~\\
\textbf{Initialization:}
\STATE Initialize particles $(\mathbf{s}^k, \mathbf{x}^l)$ where $k=1,\cdots,K,~l=1,\cdots,L$;
\STATE Initialize weights of style samples, $W_{s^k}=1/K$;
\STATE Initialize weights of viewpoint samples, $W_{x^l}=1/L$;\\
\textbf{Iteration:}
\FOR{$i=1$; $i<IterNo$; $i++$}
\STATE Compute the likelihood of particles $w_{kl}=exp{\frac{-||\mathbf{y}- \mathcal{A} \times_2 \mathbf{s}^k \times_3 \psi(\mathbf{x}^l)||^2}{ 2 \sigma^2}}$;
\STATE Update the weights of style samples $W_{s^k} =\frac{\sum^L_{l=1} w_{kl} }{ \sum^K_{k=1} \sum^L_{l=1} w_{kl}}$;
\STATE Update the weights of viewpoint samples $W_{x^l} = \frac{\sum^K_{k=1} w_{kl}}{ \sum^K_{k=1} \sum^L_{l=1} w_{kl}}$;
\STATE Keep the particle $(\mathbf{s}^*, \mathbf{x}^*)=\arg\max_{k=1,\cdots,K,~l=1,\cdots,L} \{w_{kl}\}$;
\STATE Resample $\mathbf{s}^k$ and $\mathbf{x}^l$ according to $W_{s^k}$ and $W_{x^l}$ respectively;
\ENDFOR
\RETURN $(\mathbf{s}^*, \mathbf{x}^*)$;
\end{algorithmic}
\end{algorithm}

\subsection{Multimodal Fusion}
For each individual channel (e.g. RGB and depth), a homeomorphic manifold generative model is built. Our model can be extended to include multiple modalities of information as long as there is smooth variation along the manifold as the viewpoint/pose changes.

We combine visual information (i.e. RGB) and depth information by using a combined objective function that encompasses the reconstruction error in each mapping. This is done by running the training separately on each channel and combining the objective functions. The combined reconstruction error becomes:
\begin{equation}
\label{rec_error_rgbd}
\begin{split}
E_{rgbd}(\mathbf{s}_{rgb},\mathbf{s}_d,\mathbf{x}) =
\lambda_{rgb} || \mathbf{y}_{rgb} - \mathcal{A}_{rgb}\times_2 \mathbf{s}_{rgb} \times_3 \psi(\mathbf{x})||^2  \\
+   \lambda_d || \mathbf{y}_d - \mathcal{A}_d  \times_2 \mathbf{s}_d \times_3  \psi(\mathbf{x})||^2
\end{split}
\end{equation}
Notice that the two terms share the same viewpoint variable $\mathbf{x}$. $\lambda_{rgb}$ and $\lambda_d$ were selected empirically. Since visual data has less noise than depth (which commonly exhibits missing depth values, i.e. holes), we bias the visual reconstruction error term of Eq. \ref{rec_error_rgbd}. When resampling style and viewpoint samples in our approach (Algorithm~\ref{alg:pf}), we calculate the likelihood of a particle $(\mathbf{s}_{rgb}^k, \mathbf{s}_d^k, \mathbf{x}^l)$ as
\begin{equation}
\label{eq:w_rgbd}
w_{kl} = exp{\frac{-E_{rgbd}(\mathbf{s}_{rgb}^k,\mathbf{s}_d^k,\mathbf{x}^l)}{ 2 \sigma^2}},
\end{equation}
which is a little different from Eq.~\ref{eq:w_lik}. The formulations of weights of style and viewpoint samples are the same as~\ref{eq:weight}, where visual style sample $\mathbf{s}_{rgb}^k$ and depth style sample $\mathbf{s}_d^k$ share the same weight $W_{s^k}$. This means we use a combined style sample as $\mathbf{s}^k=(\mathbf{s}_{rgb}^k,\mathbf{s}_d^k)$ for inferring. When the optimal solution $(\mathbf{s}_{rgb}^*,\mathbf{s}_d^*,\mathbf{x}^*)=\arg \min_{\mathbf{s}_{rgb},\mathbf{s}_d,\mathbf{x}} E_{rgbd}(\mathbf{s}_{rgb},\mathbf{s}_d,\mathbf{x})$ is obtained, a combined parameters $\mathbf{s}^*=[\lambda_{rgb} \mathbf{s}_{rgb}^*; \lambda_d \mathbf{s}_d^*]$ can be used for category and instance recognition.

\section{Experiments and Results}
\label{Sec:exp}


\subsection{Datasets}
\begin{figure}
  \centering
  \includegraphics[width=1.0\linewidth]{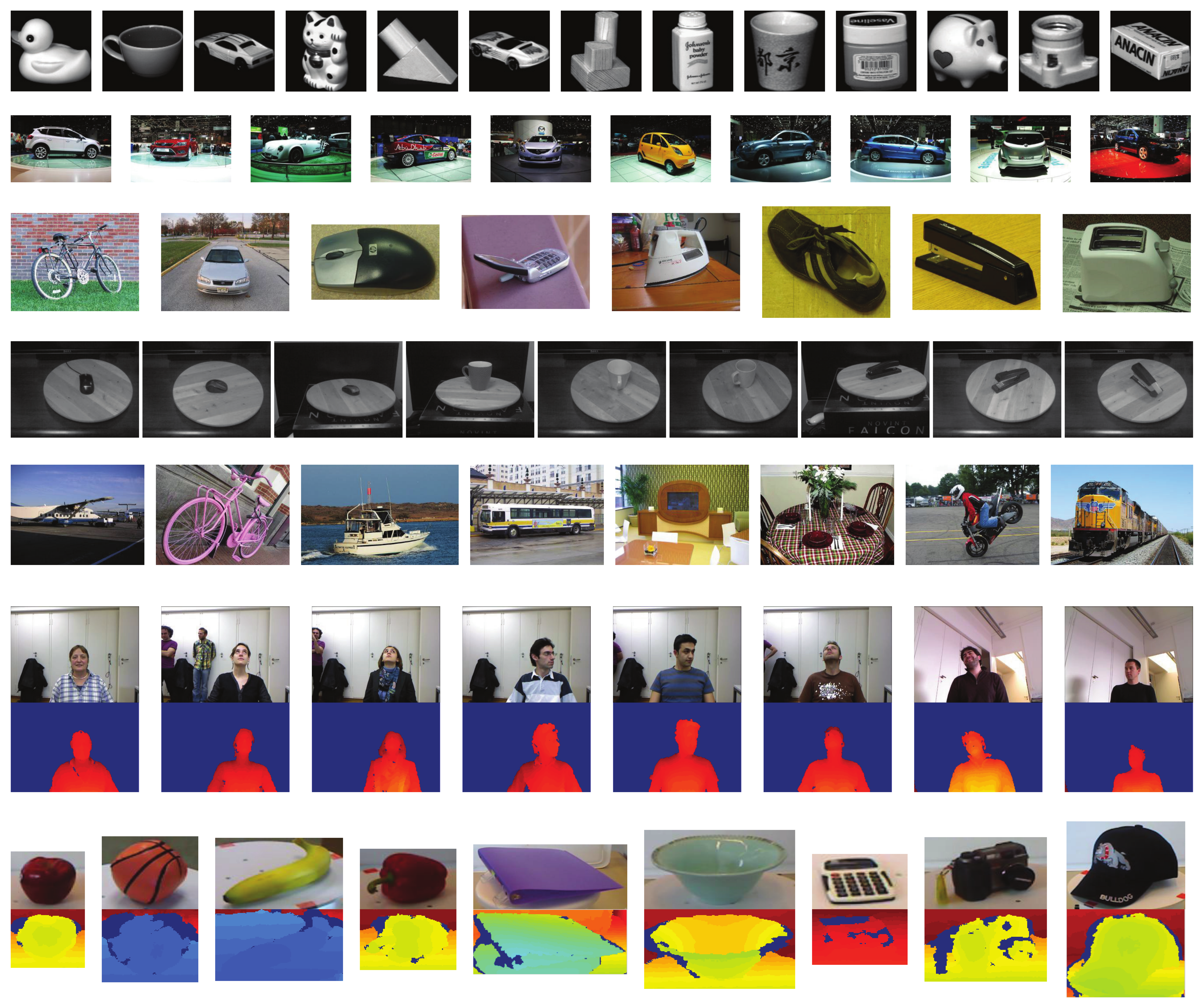}
  \caption{Sample images of different datasets. Rows from top to bottom: COIL-20 dataset~\cite{COIL20}, Multi-View Car dataset \cite{Ozuysal2009pose}, 3D Object Category dataset~\cite{Savarese07}, Table-top Object dataset~\cite{sun2010depth}, PASCAL3D+ dataset~\cite{Yu2014wacv}, Biwi Head Pose database~\cite{Fanelli2011headpose}, and RGB-D Object dataset \cite{Lai2011rgbddataset}.}
  \label{fig:sampleimage}
\end{figure}

To validate our approach we experimented on several challenging datasets: COIL-20 dataset~\cite{COIL20}, Multi-View Car dataset \cite{Ozuysal2009pose}, 3D Object Category dataset~\cite{Savarese07}, Table-top Object dataset~\cite{sun2010depth}, PASCAL3D+ dataset~\cite{Yu2014wacv}, Biwi Head Pose database~\cite{Fanelli2011headpose}, and RGB-D Object dataset \cite{Lai2011rgbddataset}. We give a brief introduction of these datasets in the following subsections. Fig.~\ref{fig:sampleimage} shows sample images from each dataset.

COIL-20 dataset~\cite{COIL20}: Columbia Object Image Library (COIL-20) dataset contains 20 objects, and each of them has 72 images captured every 5 degrees along a viewing circle. All images consist of the smallest patch (of size $128\times 128$ ) that contains the object, i.e. the background has been discarded. We used this datset for the task of arbitrary view image synthesis in order to illustrate the generative nature of our model.

Multi-View Car dataset~\cite{Ozuysal2009pose}: The Multi-View Car dataset contains 20 sequences of cars captured as the cars rotate on a rotating platform at a motor show. The sequences capture full 360 degrees images around each car. Images have been captured at a constant distance from the cars. There is one image approximately every 3-4 degrees. Finely discretized viewpoint ground truth can be calculated by using the time of capture information from the images. This dataset is suitable for the validation of dense pose estimation.

3D Object Category dataset~\cite{Savarese07}: This dataset consists of 8 object categories (bike, shoe, car, iron, mouse, cellphone, stapler and toaster). For each object category, there are images of 10 individual object instances under 8 viewing angles, 3 heights and 3 scales, \emph{i.e.} 24 different poses for each object. There are about 7000 images in total. Mask outlines for each object in the dataset are provided as well. The entire dataset can be used for multi-view object categorization and pose estimation. The car subset of 3D Object Category is typically used to evaluate the performance of sparse pose estimation.

Table-top Object dataset~\cite{sun2010depth}: This dataset contains table-top object categories with both annotated 2D image and 3D point clouds. There are two subsets called Table-Top-Local and Table-Top-Pose. Table-Top-Local subset is specific to the task of object detection and localization. We only use Table-Top-Pose subset for pose estimation task. Table-Top-Pose contains 480 images of 10 object instances for each object categories (mice, mugs and staplers), where each object instance is captured under 16 different poses (8 angles and 2 heights). Data includes the images, object masks, annotated object categories, annotated object viewpoints and 3D point clouds of the scene.

PASCAL3D+ dataset~\cite{Yu2014wacv}: PASCAL3D+ is a novel and challenging dataset for 3D object detection and pose estimation. It contains 12 rigid categories and more than 3000 object instances per category on average. The PASCAL3D+ images captured in real-world scenarios exhibit much more variability compared to the existing 3D datasets, and are suitable to test real-world pose estimation performance.

Biwi Head Pose database~\cite{Fanelli2011headpose}: This database contains 24 sequences of 20 different people (some recorded twice), captured with a Kinect sensor\footnote{http://www.xbox.com/en-us/kinect}. The subjects were recorded at roughly one meter distance to the sensor. The subjects move their heads around to try and span all possible yaw/pitch angles they could perform. There are over 15K images in the dataset. Each frame was annotated with the center of the head in 3D and the head rotation angles (respectively pith, yaw, and roll angles) by using the automatic system. For each frame, a depth image, the corresponding RGB image, and the annotation is provided. The head pose range covers about $\pm 75^\circ$ yaw, $\pm 60^\circ$ pitch, and $\pm 50^\circ$ roll. It is a good choice to use Biwi Head Pose database for 3D head pose estimation, as it provides fine ground truth of 3D rotation angles.

RGB-D Object dataset~\cite{Lai2011rgbddataset}: This dataset is large and consists of 300 common household table-top objects. The objects are organized into 51 categories. Images in the dataset were captured using a Kinect sensor that records synchronized and aligned visual and depth images. Each object was placed on a turntable and video sequences were captured for one whole rotation. There are 3 video sequences for each object each captured at three different heights ($30^\circ$, $45^\circ$, and $60^\circ$) so that the object is viewed from different elevation angles (with respect to the horizon). The dataset provides ground truth pose information for all 300 objects. Included in the RGB-D Object dataset are 8 video sequences of common indoor environments (office workspaces, meeting rooms, and kitchen areas) annotated with objects that belong to the RGB-D Object dataset. The objects are visible from different viewpoints and distances, and may be partially or completely occluded. These scene sequences are part of the RGB-D Scenes dataset. As one of the largest and most challenging multi-modal multi-view datasets available, we used RGB-D Object dataset for joint category, instance and pose estimation on multi-modal data, and used it to test a near real-time system we built for category recognition of table-top objects.

\subsection{Parameter Determination}
As in Subsection~\ref{SSec:mapping}, there is one key parameter in our model that significantly affects the performance. This is the number of mapping centers $M$. This parameter determines the density of arbitrary points $\mathbf{z}_j$ on the homeomorphic manifold when learning the nonlinear mapping function. If $M$ is too small, the learnt mapping function may be not able to model the relationship between view manifolds and visual inputs well enough. On the other hand, the computation cost of learning the mapping function will increase in proportion to $M$. When $M$ is larger than the number of training data points the learning problem becomes ill-posed. In addition to $M$, the image features are also important for our model. The images features are what represent the objects in the visual/input space. However our approach is orthogonal to the choice of the image representation, and any vectorized representation can be used.

\begin{figure}
  \centering
  \includegraphics[width=0.8\linewidth]{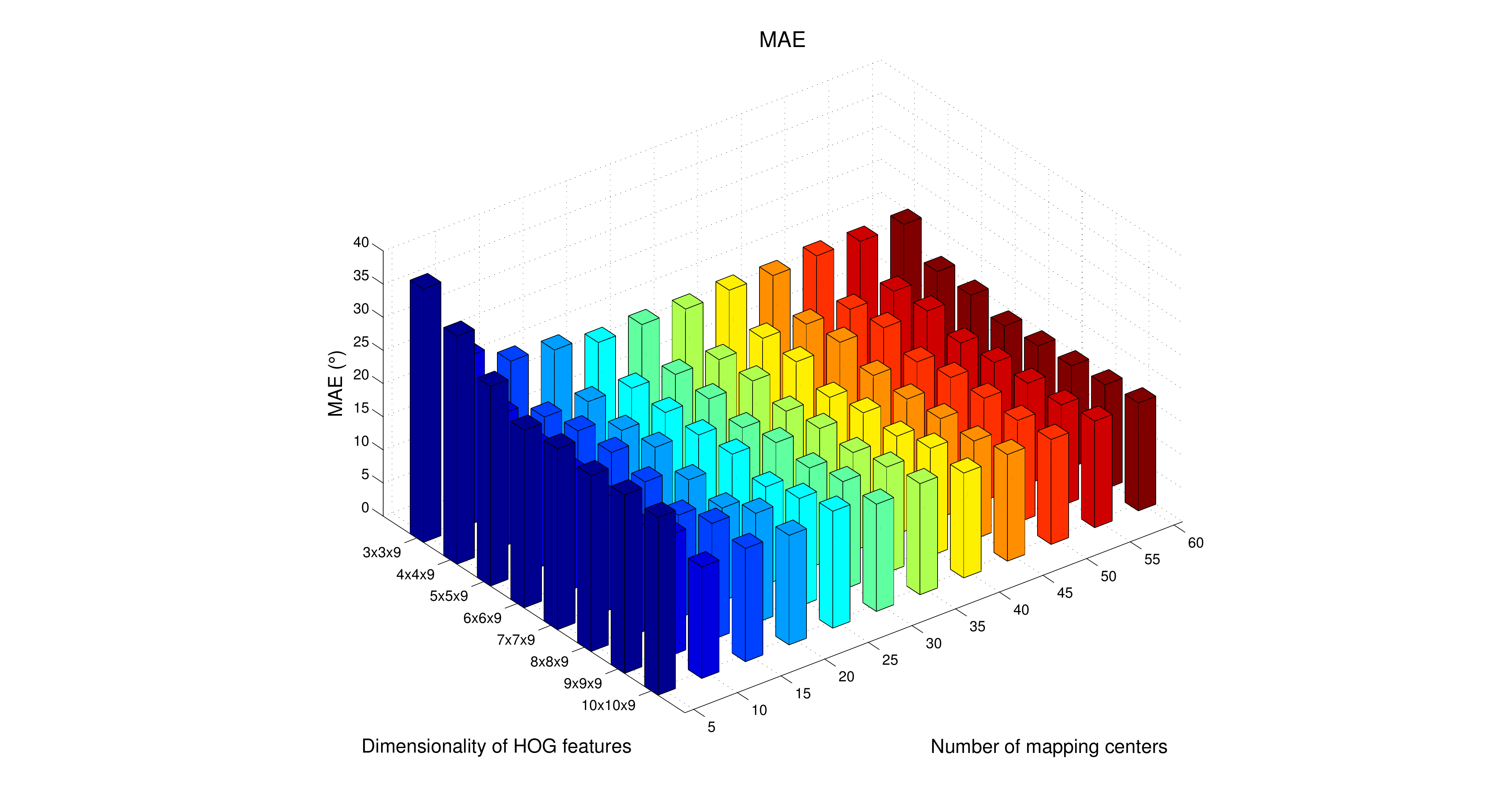}
  \includegraphics[width=0.8\linewidth]{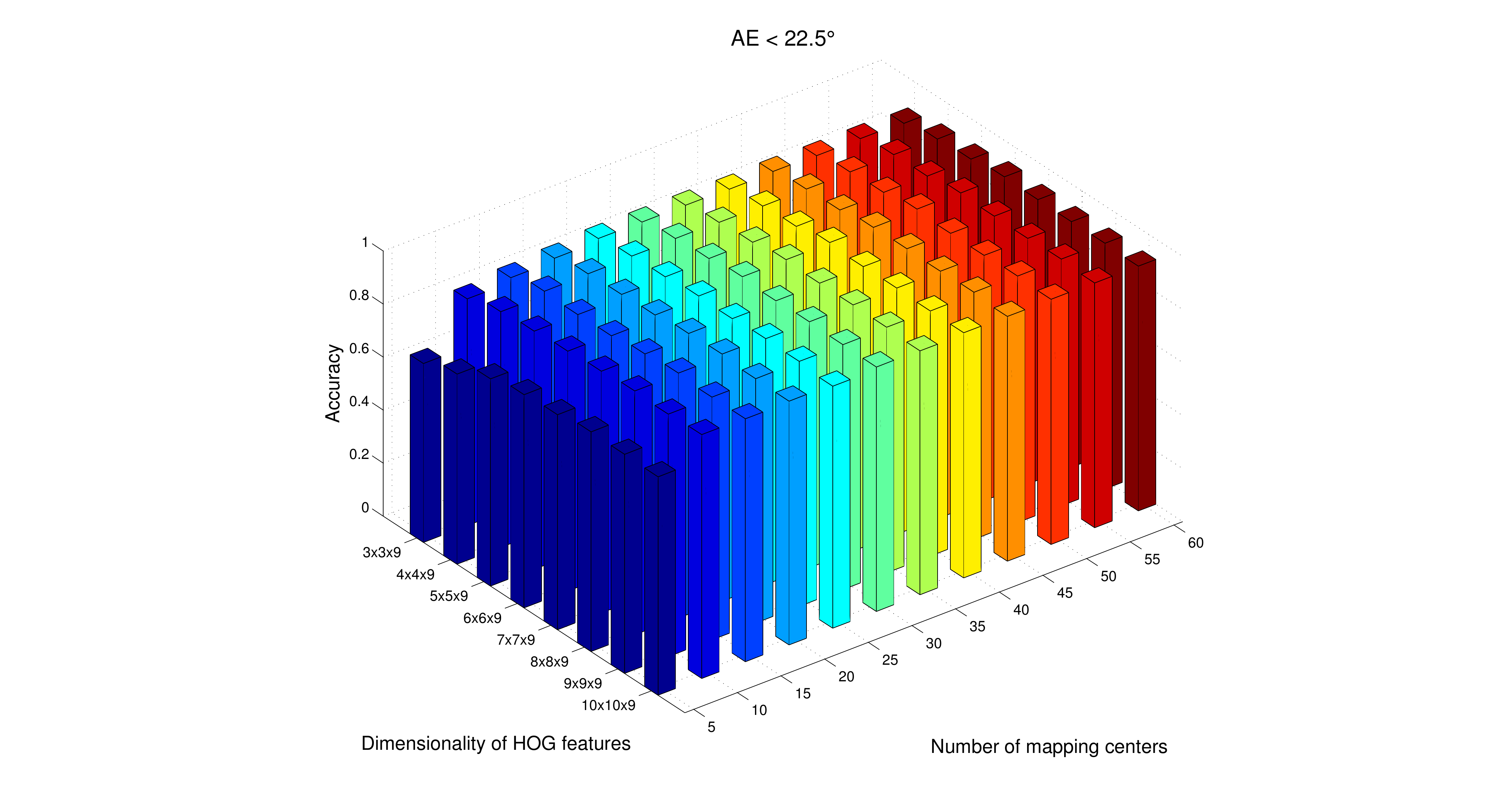}
  \includegraphics[width=0.8\linewidth]{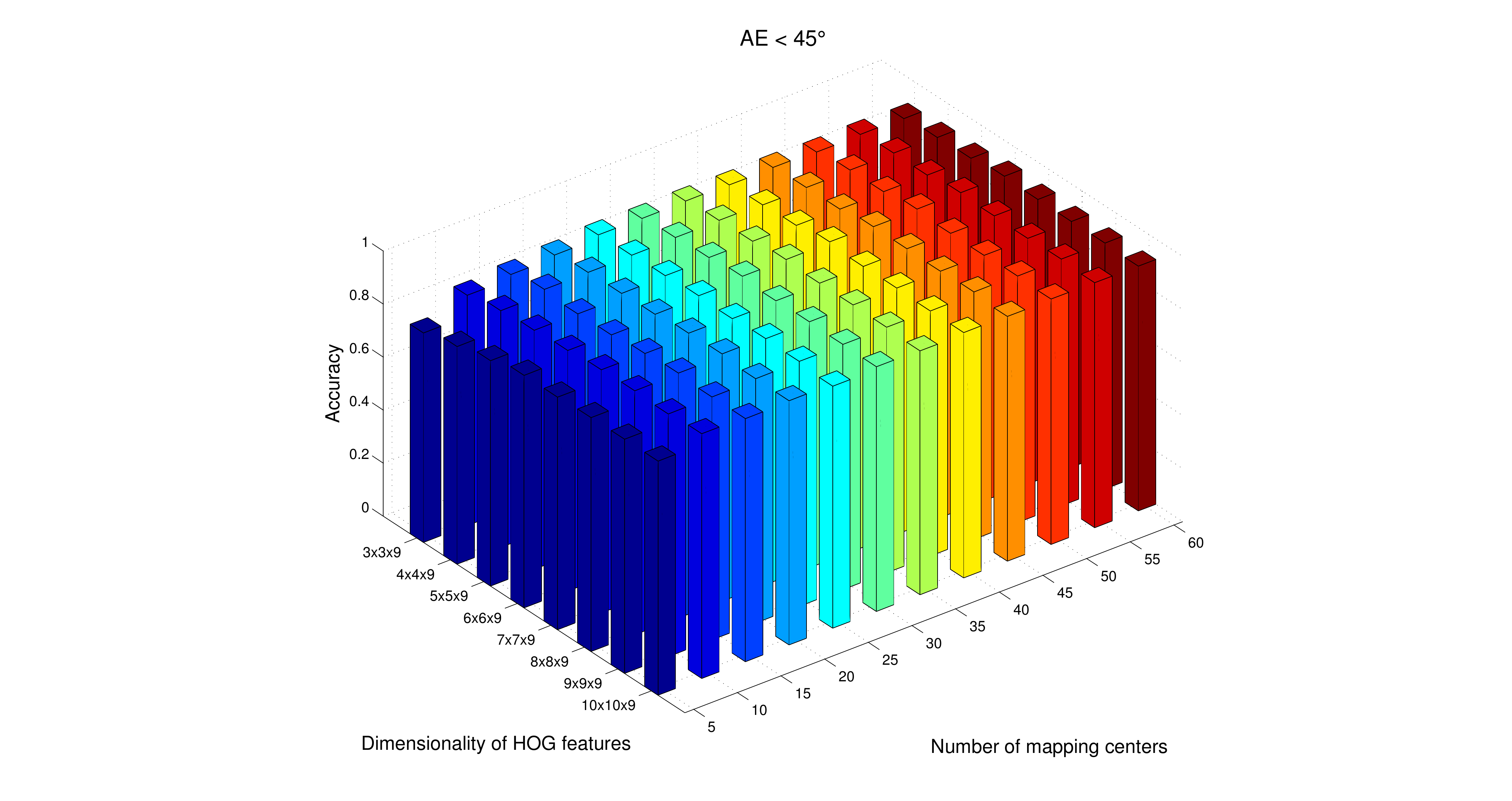}
  \caption{Cross validation results on Multi-View Car dataset for parameter determination. x and y axes are the number of mapping centers and the dimensionality of HOG features, z axis is the pose estimation performance. Titles are shown on the axes (zoom in to see the text). From top to bottom: MAE, $AE<22.5^\circ$, and $AE<45^\circ$. The dimensionality of HOG features is indicated as $n\times n\times 9$, meaning that HOG features are computed in $n \times n$ grids with 9 orientation bins.}
  \label{fig:cross_vali}
\end{figure}

To get proper parameters, we performed cross validation within the training data of each fold. For example, in the 50\% split experiment of Subsection~\ref{SSec:20cars}, we learnt our model on 9 out of the 10 car sequences in the training set and tested using the 1 left out. We performed 10 rounds of cross validation. Fig.~\ref{fig:cross_vali} shows the performance of our model with different parameters: the dimensionality of HOG~\cite{Dalal05hog} features we used, and the number of mapping center ($M$). We used 35 mapping centers along a 2D unit circle to define the kernel map $\psi(\cdot)$ in Eq~\ref{eq:learned_style_model}, and used HOG features calculated in $7 \times 7$ grids with 9 orientation bins to represent the inputs. The results in Table~\ref{tab:cars} were obtained using these parameters. Such cross validation is performed for each experiment in this section.

\subsection{Arbitrary View Synthesis}
\label{SSec:viewsyn}

\begin{figure}
  \centering
  \includegraphics[width=1\linewidth]{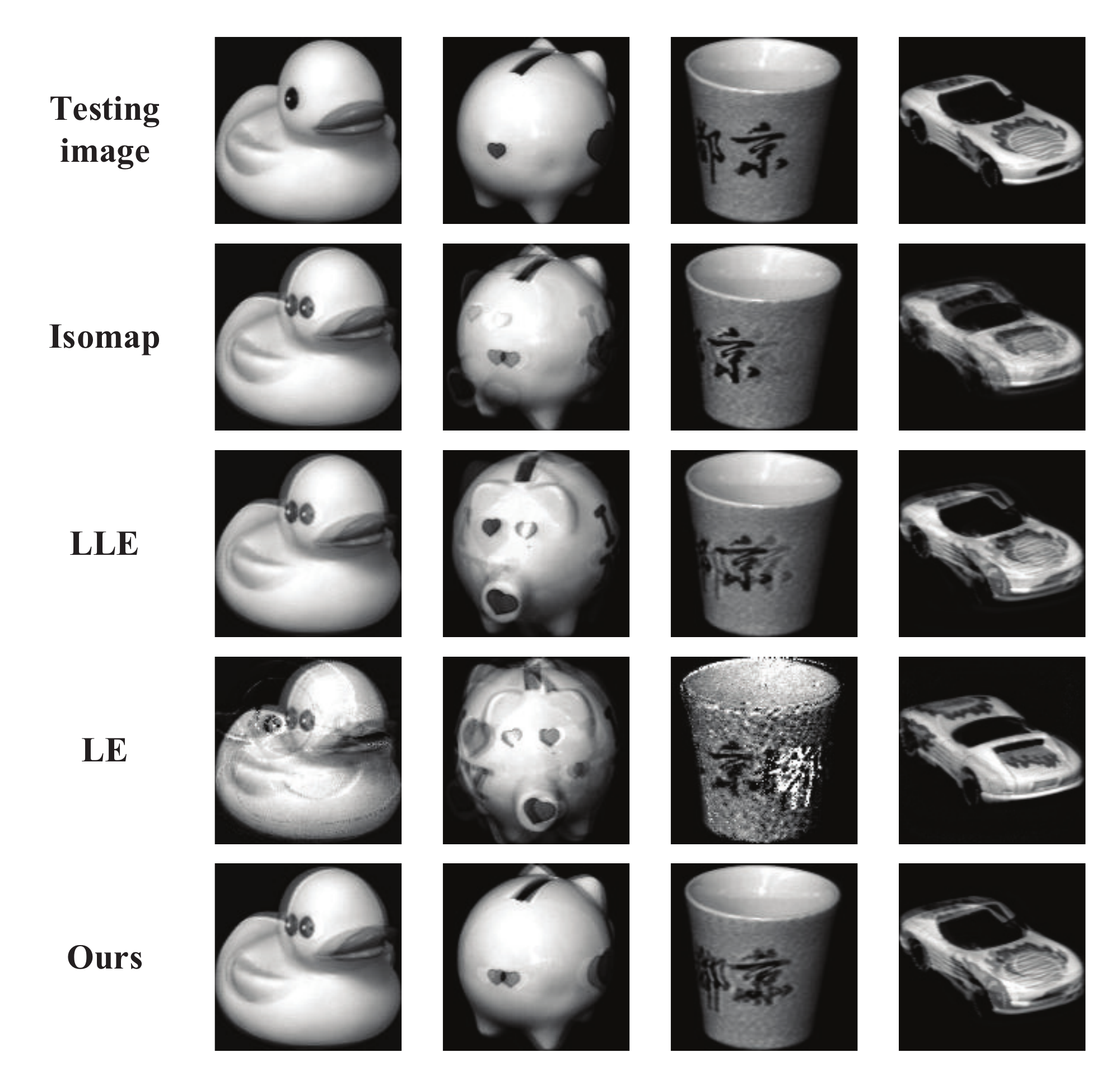}
  \caption{Synthesized images of unseen views. The first row shows image samples in testing set, and the rest four rows show synthesized images. Our results are visually better than other manifold learning methods, and are more robust as well.}
  \label{fig:viewsyn}
\end{figure}

Since our model is a generative model mapping from the manifold representation to visual inputs, we can perform arbitrary view synthesis if image intensities are used as visual inputs. We did arbitrary view synthesis experiments on COIL-20 dataset to show the generative nature of our model. We used 54 images to learn our generative model for each object in COIL-20 dataset, and tested the rest 18 images (every 4th), i.e. synthesized images from the viewpoints of the 18 testing images. We report mean squared error (MSE) to evaluate the synthesized images, which can be defined as following
\begin{equation}
MSE=\frac{1}{MN}\sum^M_{i=1}\sum^N_{j=1}[I_o(i,j)-I_s(i,j)]^2
\end{equation}
where $I_o(i,j)$ is the intensity of the pixel located at $(i,j)$ in the testing image $I_o$ of size $M\times N$, and $I_s$ is the synthesized image from the same viewpoint of image $I_o$. For comparison, we also used typical manifold learning methods, including LLE~\cite{Roweis00}, Isomap~\cite{Tenenbaum98}, and Laplacian Eigenmap (LE)~\cite{Belkin03NC}, to learn a latent representing of the view manifold, and then learned a similar generative map as Eq.~\ref{eq:wholemap}. Notice that different from the conceptual manifold used in Subsection~\ref{SSec:viewmani} where the embedded coordinates can be computed according to Eq~\ref{eq:mani} given the pose angles, the embedding coordinates of the unseen views (in the testing set) are obtained by linear interpolation between its neighbors in the training set with the assumption that the manifold learned by LLE, Isomap, or LE is locally linear. Results in Table~\ref{tab:result_viewsyn} and Fig.~\ref{fig:viewsyn} show that our model can correctly generate unseen view of a learned object, and our synthesis results are both quantitatively and qualitatively better than those obtained by typical manifold learning methods.

\begin{table}[htb]
\caption{Arbitrary view synthesis results on COIL-20 dataset}
\label{tab:result_viewsyn}
\begin{center}
\begin{tabular}{|l|c|}
\hline Method & Mean Squared Error\\
\hline
\hline Isomap~\cite{Tenenbaum98} & 704\\
\hline LLE~\cite{Roweis00} & 950\\
\hline LE~\cite{Belkin03NC} & 7033\\
\hline Ours & \textbf{361}\\
\hline
\end{tabular}
\end{center}
\end{table}

\subsection{Dense Viewpoint Estimation}
\label{SSec:20cars}
We experimented on the Multi-View Car dataset to evaluate our model for dense pose estimation. Following previous approaches ~\cite{Ozuysal2009pose,Torki11}, there are two experimental setups: \emph{50\% split} and \emph{leave-one-out}. For the former we take the first 10 cars for training and the rest for testing, resulting a 10-dimensional style space. For the latter we learn on 19 cars and test on the remaining 1, and the dimensionality of the style space is 19. Pixels within the bounding box (BBox) provided by the dataset were used as inputs.

For quantitative evaluation, we use the same evaluation criterion as~\cite{Ozuysal2009pose,Torki11}, \emph{i.e.} Mean Absolute Error (MAE) between estimated and ground truth viewpoints. To compare with classification-based viewpoint estimation approaches (which use discrete bins) we also compute the percentages of test samples that satisfy $AE<22.5^{\circ}$ and $AE<45^{\circ}$ where the Absolute Error (AE) is $AE= | Estimated Angle - Ground Truth|$). According to~\cite{Torki11}, the percentage accuracy in terms of $AE<22.5^{\circ}$ and $AE<45^{\circ}$ can achieve equivalent comparison with classification-based pose estimation approaches that use 16-bin and 8-bin viewpoint classifiers respectively.

We represented the input using HOG features. Table~\ref{tab:cars} shows the view estimation results in comparison to the state-of-the-art. Notice that results of~\cite{Torki11} were achieved given bounding boxes of the cars while those of~\cite{Ozuysal2009pose} were without bounding boxes, i.e. simultaneously performed localization. The quantitative evaluation clearly demonstrates the significant improvement we achieve.

\begin{table*}
\caption{Results on Multi-View Car dataset}
\label{tab:cars}
\begin{center}
\begin{tabular}{|l|c|c|c|}
\hline
Method & MAE ($^\circ$) &  \% of $AE<22.5^\circ$  &  \% of $AE<45^\circ$  \\
\hline
\hline \cite{Ozuysal2009pose} & 46.48 & 41.69 & 71.20 \\
\hline \cite{Torki11} - leave-one-out & 35.87 & 63.73 & 76.84 \\
\hline \cite{Torki11} - 50\% split & 33.98 & 70.31 & 80.75 \\
\hline \textbf{Ours - leave-one-out} & \textbf{19.34} & \textbf{90.34} & \textbf{90.69} \\
\hline \textbf{Ours - 50\% split} & \textbf{24.00} & \textbf{87.77} & \textbf{88.48} \\
\hline
\end{tabular}
\end{center}
\end{table*}

\subsection{Sparse Pose Estimation}
To validate our approach for viewpoint estimation using sparse training samples on the viewing circle we did experiments on the 3D Object Category dataset and the Table-top Object dataset. Both datasets contain only 8 sparse views on the viewing circle. We used HOG features as input, calculated within the BBoxes (obtained from the mask outlines).

For 3D Object Category dataset, we used its car subset and bicycle subset, and followed the same setup as ~\cite{Savarese07,SavareseFeiFei2009}: 5 training sequences and 5 sequences for testing (160 training and 160 testing images). The Table-Top-Pose subset of the Table-top dataset was used for evaluating the viewpoint estimation of the following classes: staplers, mugs and computer mice. We followed the same setup as ~\cite{sun2010depth,Payet2011}: the first 5 object instances are selected for training, and the remaining 5 for testing. Following the above setups, the dimensionality of the style space we used are both 5.

For comparison with~\cite{Savarese07,SavareseFeiFei2009,schmid20103Dgeometricmodels,Payet2011,Torki11,sun2010depth}, we report our results in terms of $AE<45^\circ$ (equivalent to an 8-bin classifier). Results are shown in Table~\ref{tab:result_sparse}. Some of the state-of-the-art algorithms mentioned to jointly do detection and pose estimation (without BBox) and reported pose estimation only for successfully detected objects, while we do pose estimation for all objects in the database given BBoxes. Therefore, the comparisons in Table~\ref{tab:result_sparse} may be not completely fair. We indicate the setting for each approach and put results in the corresponding columns in Table~\ref{tab:result_sparse}. As shown in Table~\ref{tab:result_sparse}, our homeomorphic manifold analysis framework achieves 93.13\% on the car subset of the 3D Objects dataset. This is far more than the state-of-the-art result of 85.38\% in~\cite{Payet2011} and 77.5\% in~\cite{Torki11}. On the bicycle subset of the 3D Objects dataset our accuracy is 94.58\%. This is more than 17\% and 25\% improvement over the results in~\cite{Payet2011} and~\cite{schmid20103Dgeometricmodels}, respectively. We also achieve the best average accuracy of 89.17\% on the three classes of the Table-Top-Pose subset, improving about 26\% and 43\% over ~\cite{Payet2011} and~\cite{sun2010depth}, respectively. These results show the ability of our framework to model the visual manifold, even with sparse views.

\begin{table*}[htb]
\caption{Sparse pose estimation results and comparison with the state-of-the-arts}
\label{tab:result_sparse}
\begin{center}
\small{
\begin{tabular}{|l|l|c|c|}
\hline
Dataset & Method & Pose estimation & Pose estimation\\
 &  & (without BBox) & (with BBox)\\
\hline
\hline 3D Object Category (car)& \cite{Savarese07} & 52.5\% & -\\
\hline 3D Object Category (car)& \cite{SavareseFeiFei2009} & 66.63\% & -\\
\hline 3D Object Category (car)& \cite{schmid20103Dgeometricmodels} & 69.88\% & -\\
\hline 3D Object Category (car)& \cite{Torki11} & - & 77.5\%\\
\hline 3D Object Category (car)& \cite{Payet2011} & - & 85.38\%\\
\hline 3D Object Category (car)& Ours & - & 93.13\%\\
\hline
\hline 3D Object Category (bike)& \cite{schmid20103Dgeometricmodels} & 75.5\% & -\\
\hline 3D Object Category (bike)& \cite{Payet2011} & - & 80.75\%\\
\hline 3D Object Category (bike)& Ours & - & 94.58\%\\
\hline
\hline Table-Top-Pose& \cite{sun2010depth} & - & 62.25\%\\
\hline Table-Top-Pose& \cite{Payet2011} & - & 70.75\%\\
\hline Table-Top-Pose& Ours & - & 89.17\%\\
\hline
\end{tabular}
}
\end{center}
\end{table*}


\subsection{Pose Estimation on PASCAL3D+ dataset}

We performed pose estimation on PASCAL3D+ dataset~\cite{Yu2014wacv}. Such a novel and challenging dataset is suitable to test pose estimation performance in real-world scenarios. We also used HOG features calculated within the BBoxes as input. We tested our model on 11 categories as the benchmark~\cite{Yu2014wacv}, including aeroplane, bicycle, boat, bus, car, chair, dining table, motorbike, sofa, train, and tv monitor, following the same experimental setting as \cite{Yu2014wacv}. Results in Table~\ref{tab:pascal3d} show the power of our model for pose estimation. Noting that the benchmark results of \cite{Yu2014wacv} were performed simultaneously with detection, the comparison in Table~\ref{tab:pascal3d} is not completely fair.

\begin{table}[htb]
\caption{Pose performance (\%) on PASCAL3D+ dataset. It can be seen that our approach outperforms VPDM \cite{Yu2014wacv}}
\label{tab:pascal3d}
\begin{center}
\small{
\begin{tabular}{|l||c|c||c|c|}
\hline
Class & \scriptsize{Ours (\% of $AE<45^\circ$)}  & \scriptsize{VDPM-8V} & \scriptsize{Ours (\% of $AE<22.5^\circ$)} & \scriptsize{VDPM-16V}\\
\hline
\hline
aeroplane & 60.3 & 23.4 & 40.2 & 15.4\\
\hline
bicycle & 60.7 & 36.5 & 40.3 & 18.4\\
\hline
boat & 39.7 & 1.0 & 20.6 & 0.5\\
\hline
bus & 73.0 & 35.5 & 68.7 & 46.9\\
\hline
car & 55.4 & 23.5 & 46.4 & 18.1\\
\hline
chair & 50.0 & 5.8 & 34.1 & 6.0\\
\hline
diningtable & 45.2 & 3.6 & 37.5 & 2.2\\
\hline
motorbike & 67.2 & 25.1 & 48.9 & 16.1\\
\hline
sofa & 75.9 & 12.5 & 59.2 & 10.0\\
\hline
train & 56.0 & 10.9 & 48.0 & 22.1\\
\hline
tvmonitor & 80.1 & 27.4 & 55.1 & 16.3\\
\hline
\hline
average & 59.0 & 18.7 & 44.2 & 15.6\\
\hline
\end{tabular}
}
\end{center}
\end{table}

\subsection{3D Head Pose Estimation}

To test our model for multi-view object pose estimation with 3D pose/viewpoint variation we performed experiments on the Biwi Head Pose database ~\cite{Fanelli2011headpose}. In our experiments, we only considered the problem of pose estimation and not head detection. We assumed that the faces were detected successfully, thus we just used the depth data within the bounding boxes (obtained from the provided masks) to compute HOG features. Head poses were represented on a 3-dimensional conceptual manifold in 4D Euclidean space, \emph{i.e.} a normalized 3-sphere. For comparison, we ran a 5-fold and a 4-fold subject-independent cross validation on the entire dataset, resulting a 16-dimensional and a 15-dimensional style space respectively. This is the same experimental setup as~\cite{Fanelli2011headpose} and ~\cite{Fanelli2013randomforests}. We also reported the mean and standard deviation of the errors for each rotation angles. Results are shown in Table~\ref{tab:biwi}. It should be noticed that the pose results of~\cite{Fanelli2011headpose} and ~\cite{Fanelli2013randomforests} in Table~\ref{tab:biwi} are computed only for correctly detected heads with 1.0\% and 6.6\% missed respectively. It can been seen that our model significantly outperforms ~\cite{Fanelli2011headpose} in 5-fold cross validation. In 4-fold cross validation, our mean errors are a little higher than~\cite{Fanelli2013randomforests} with respect to yaw and pitch, but our standard deviations are lower, which means that our estimation results are more stable. These results show the ability of our model to solve continuous 3D pose estimation robustly.

\begin{table*}[htb]
\caption{Summary of results on Biwi Head Pose database}
\label{tab:biwi}
\begin{center}
\small{
\begin{tabular}{|l|c|c|c|c|}
\hline
Method & Validation & Yaw error & Pitch error & Roll error\\
\hline
\hline Ours (Depth)& 5-fold & $4.72\pm4.69^\circ$ & $3.84\pm3.90^\circ$ & $4.78\pm5.49^\circ$\\
\hline Ours (RGB)& 5-fold & $8.09\pm7.90^\circ$ & $6.46\pm6.79^\circ$ & $6.00\pm6.49^\circ$\\
\hline \textbf{Ours (RGB+D)}& 5-fold & $\mathbf{4.67\pm4.57^\circ}$ & $\mathbf{3.85\pm3.68^\circ}$ & $\mathbf{4.59\pm5.24^\circ}$\\
\hline
Baseline~\cite{Fanelli2011headpose} (Depth)& 5-fold & $9.2\pm13.7^\circ$ & $8.5\pm10.1^\circ$ & $8.0\pm8.3^\circ$\\
\hline
\hline Ours (Depth)& 4-fold & $4.84\pm4.78^\circ$ & $3.87\pm4.06^\circ$ & $4.79\pm5.61^\circ$\\
\hline Ours (RGB)& 4-fold & $8.40\pm8.31^\circ$ & $6.60\pm6.87^\circ$ & $6.10\pm6.65^\circ$\\
\hline Ours (RGB+D)& 4-fold & $4.81\pm4.77^\circ$ & $3.86\pm3.93^\circ$ & $\mathbf{4.73\pm5.49^\circ}$\\
\hline
Baseline~\cite{Fanelli2013randomforests} (Depth)& 4-fold & $3.8\pm6.5^\circ$ & $3.5\pm5.8^\circ$ & $5.4\pm6.0^\circ$\\
\hline
\end{tabular}
}
\end{center}
\end{table*}

\subsection{Categorization and Pose Estimation}

We used the entire 3D Objects Category dataset to evaluate the performance of our framework on both object categorization and viewpoint estimation. Similar to~\cite{Savarese07,SavareseFeiFei2009}, we tested our model on an 8-category classification task, and the farthest scale is not considered. We followed the same experimental setting as~\cite{Savarese07} by randomly selecting 8/10 object instances for learning and the remaining 2 instances for testing. Since there are totally 64 instances for training, the dimensionality of the style space used in this experiment is 64. Average recognition results of 45 rounds are shown in Table ~\ref{tab:3dobject}. We achieve an average recognition accuracy of 80.07\% on 8 classes and an average pose estimation performance of 73.13\%\footnote{Notice that only pose results of correctly categorized images were taken for evaluation.} on the entire test set which satisfies $AE<45^\circ$. We achieve markedly higher accuracy in recognition in 5 of the 8 classes than \cite{Savarese07}. However our performance is not better than~\cite{Pepik12}, which shows the room to improve the categorization capability of our model. In fact, a follow up paper \cite{bakry2014untangling} that uses our framework with a feed foreword solution (without sampling) achieves much better results than~\cite{Pepik12}.

\begin{table}
\caption{Category recognition performance (\%) on 3D Object Category dataset}
\label{tab:3dobject}
\begin{center}
\small{
\begin{tabular}{|l|c|c|c|}
\hline
Class & Ours & Baseline~\cite{Savarese07} & \cite{Pepik12}\\
\hline
\hline
Bicycle & 99.79 & 81.00 & 98.8\\
\hline
Car & 99.03 & 70.00 & 99.8\\
\hline
Cellphone & 66.74 & 76.00 & 62.4\\
\hline
Iron & 75.78 & 77.00 & 96.0\\
\hline
Mouse & 48.60 & 87.00 & 72.7\\
\hline
Shoe & 81.70 & 62.00 & 96.9\\
\hline
Stapler & 82.66 & 77.00 & 83.7\\
\hline
Toaster & 86.24 & 75.00 & 97.8\\
\hline
\end{tabular}
}
\end{center}
\end{table}

\subsection{Joint Object and Pose Recognition}

We evaluated our model for joint object and pose recognition on multi-modal data by using the RGB-D Object dataset. Training and testing follows the exact same procedure as \cite{Lai2011optree}. Training was performed using sequences at heights: $30^\circ$ and $60^\circ$. Testing was performed using the $45^\circ$ height sequence. We treated the images of each instance in the training set as one sequence, thus resulted a 300-dimensional style space. We used HOG features for both RGB channels and depth channel. We also experimented with an additional more recent depth descriptor called Viewpoint Feature Histogram (VFH) \cite{Rusu2010VFH} computed on the 3D point cloud data.

\begin{figure}[htb]
  \centering
  \includegraphics[width=4.3in, height=2.0in]{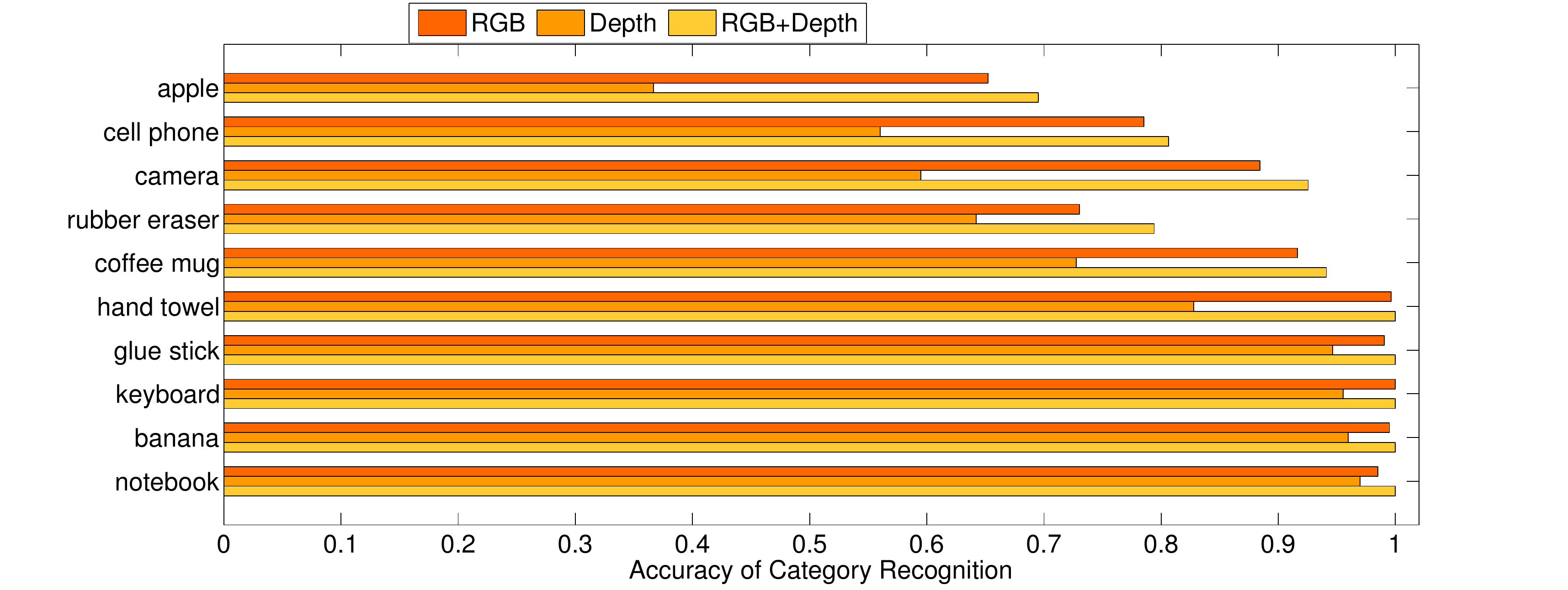}
  \centering
  \includegraphics[width=4.5in, height=2.5in]{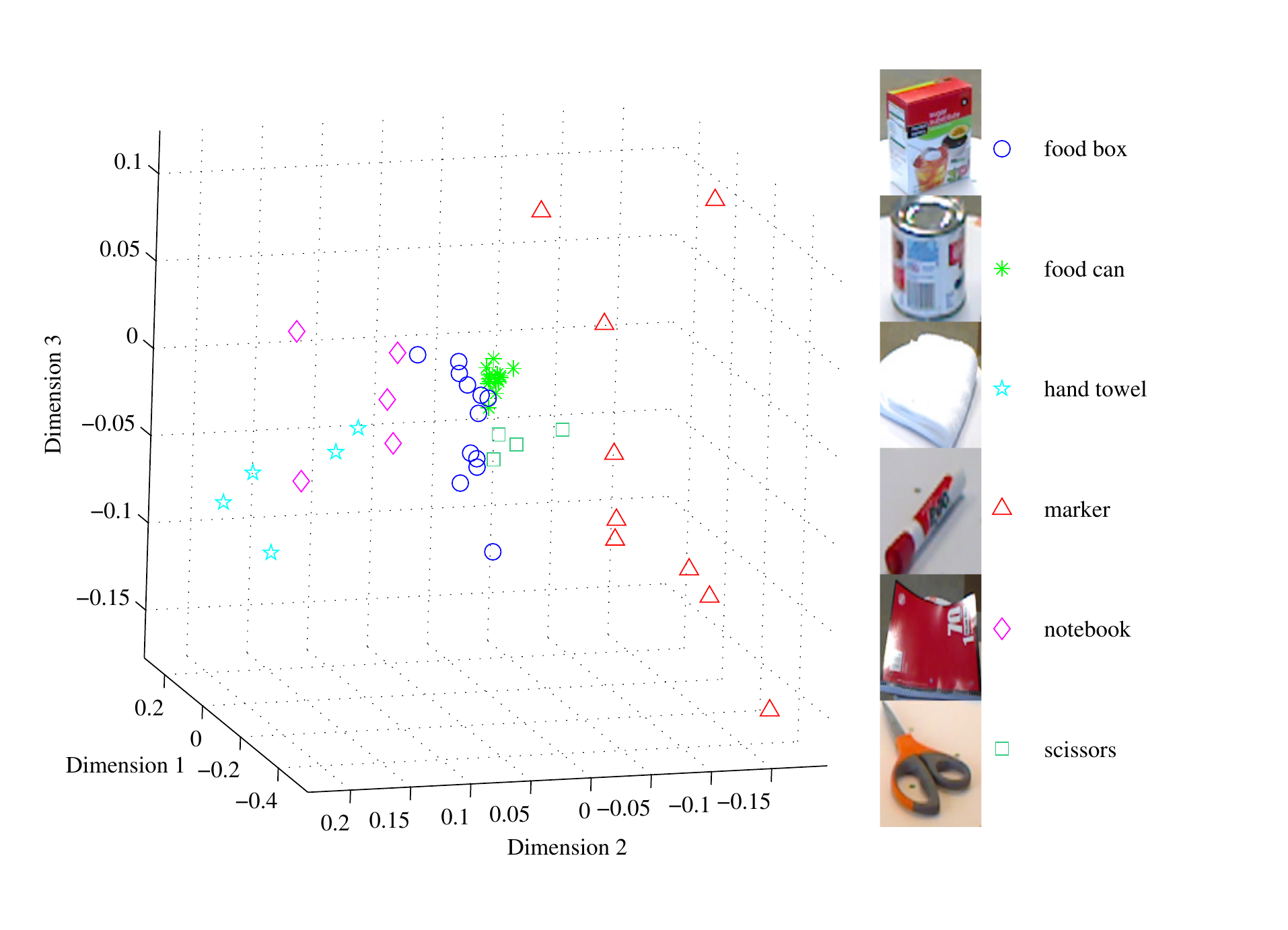}
  \caption{Top: Category recognition using different modes for a subset of categories in RGB-D Object dataset. Bottom: Sampled instances from 6 different categories in RGB-D Object dataset. Notice: flatter objects lie to the left and more rounded shapes to the right}
  \label{fig:selvect}
\end{figure}

\begin{figure}[h]
\centering
\includegraphics[width=1.0\linewidth]{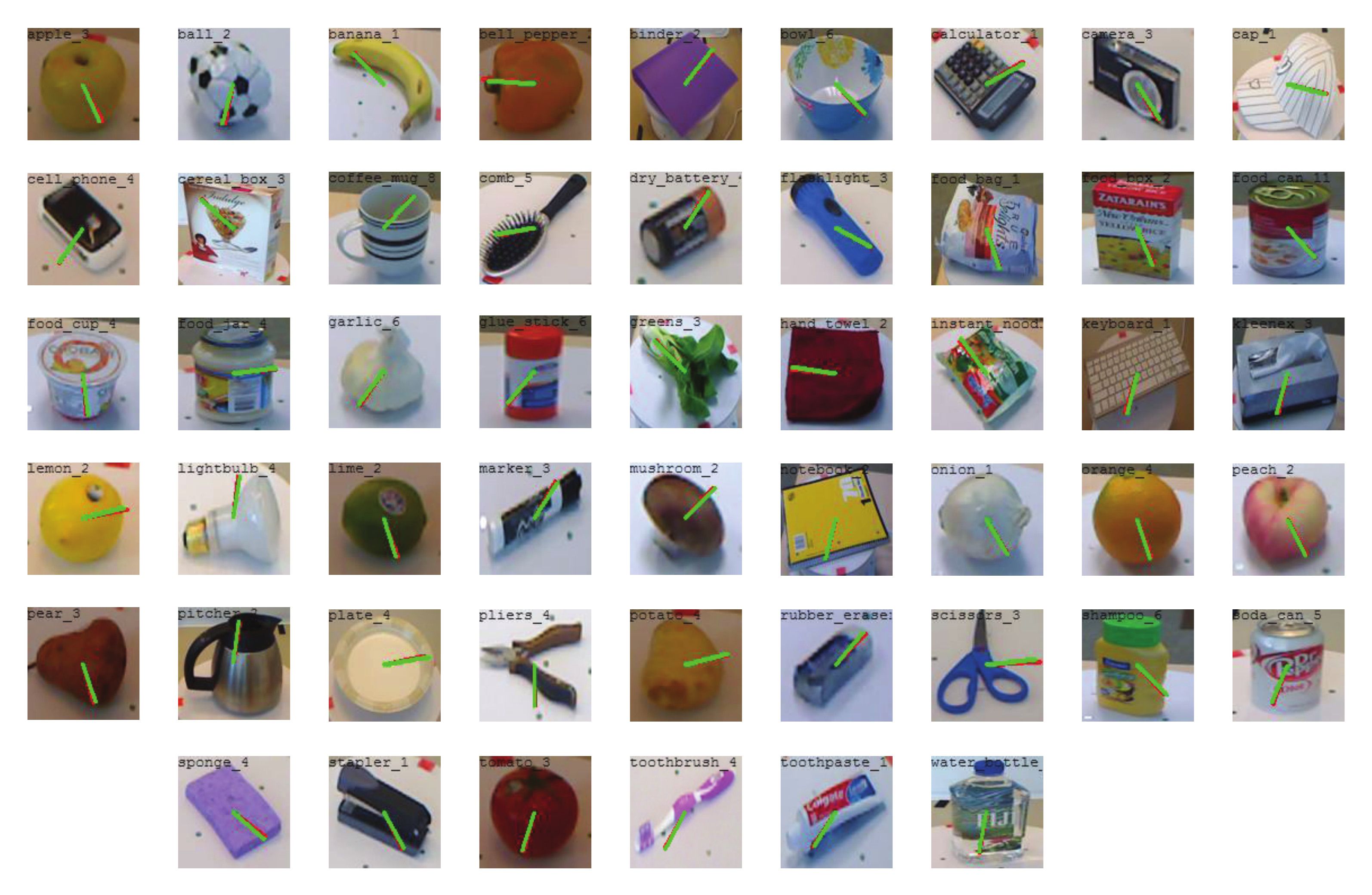}
\caption{Sample correct results for object and pose recognition on RGB-D Object dataset. Black text: category name and instance number. Red line: estimated pose. Green line: ground truth pose.}
\label{fig:result-rgbd}
\end{figure}

Table \ref{fulltable} summarizes the results of our approach and compares to 2 state-of-the-art baselines. In the case of category and instance recognition (column 2 \& 3), we achieve results on par with the state-of-the-art \cite{Lai2011optree}. We find that $\approx 57\%$ of the categories exhibit better category recognition performance when using RGB+D, as opposed to using RGB only (set of these categories shown in Fig.~\ref{fig:selvect}-top). Fig.~\ref{fig:selvect}-bottom shows an illustration of sample instances in the object style latent space. Flatter objects lie more towards the lefthand side and rounder objects lie more towards the righthand side. Sample correct results for object and pose recognition are shown in Fig.~\ref{fig:result-rgbd}.

Incorrectly classified objects were assigned pose accuracies of 0. Avg. and Med. Pose (C) are computed only on test images whose categories were correctly classified. Avg. and Med. Pose (I) were computed only using test images that had their instance correctly recognized. All the object pose estimations significantly out-performs the state-of-the-art \cite{Lai2011optree,ElGaaly2012rgbd_mkl}. This verifies that the modeling of the underlying continuous pose distribution is very important in pose recognition.

Lime and bowl categories were found to have better category recognition accuracy when using depth only instead of using either visual-only or visual and depth together. This can be explained by the complete lack of visual features on their surfaces. Some object instances were classified with higher accuracy using depth only also. There were 19 (out of 300) of these instances, including: lime, bowl, potato, apple and orange. These instances have textureless surfaces with no distinguishing visual features and so the depth information alone was able to utilize shape information to achieve higher accuracy.

In Table \ref{fulltable} we see that depth HOG (DHOG) performs quite well in all the pose estimation experiments except for where misclassified categories or instances were assigned 0 (column 3 \& 4). DHOG appears to be a simple and effective descriptor to describe noisy depth images captured by the Kinect in the dataset. It achieves better accuracy than \cite{Lai2011optree} in the pose estimation. Similar to \cite{Lai2011optree}, recursive median filters were applied to depth images to fill depth holes. This validates the modeling of the underlying continuous distribution which our homeomorphic manifold mapping takes advantage of. VFH is a feature adapted specifically to the task of viewpoint estimation from point cloud data. No prior point cloud smoothing was done to filter out depth holes and so its performance suffered.

\begin{table*}[htb]
\caption{Summary of results on RGB-D Object dataset using RGB/D and RGB+D (\%). Cat. and Inst. refer to category recognition and Instance recognition respectively}
\label{fulltable}
\begin{center}
\footnotesize{
\begin{tabular}{|c||p{0.5cm}|p{0.5cm}|p{0.6cm}|p{0.6cm}|p{0.6cm}|p{0.6cm}|p{0.6cm}|p{0.6cm}|p{0.6cm}|}
\hline Methods & Cat. & Inst. & Avg Pose & Med Pose & Avg Pose (C) & Med Pose (C) & Avg Pose (I) & Med Pose (I)\\
\hline
\hline Ours (RGB) & 92.00 & 74.36 & 61.59 & 89.46 & 80.36 & 93.50 & 82.83 & 93.90\\
\hline Linear SVM (RGB) & 75.57 & 41.50 & - & - & - & - & - & -\\
\hline Ours (Depth - DHOG) & 74.49 & 36.18 & 26.06 & 0.00 & 66.36 & 86.60 & 72.04 & 90.03\\
\hline Linear SVM (Depth - DHOG) & 65.30 & 18.50 & - & - & - & - & - & -\\
\hline Ours (Depth - VFH) & 27.88 & 13.36 & 7.99 & 0.00 & 57.79 & 62.75 & 59.82 & 67.46\\
\hline
\hline \textbf{Ours (RGB+D)} & 93.10  & 74.79 & \textbf{61.57} & \textbf{89.29} & \textbf{80.01} & \textbf{93.42} & \textbf{82.32} & \textbf{93.80}\\
\hline Baseline (RGB+D) \cite{Lai2011optree} & 94.30 & 78.40 & 53.30 & 65.20 & 56.80 & 71.40 & 68.30 & 83.20\\
\hline Linear SVM (RGB+D) & 86.86 & 47.42 & - & - & - & - & - & -\\
\hline Baseline (RGB+D) \cite{ElGaaly2012rgbd_mkl} & - & - & - & - & 74.76 & 86.70 & - & -\\
\hline
\end{tabular}
}
\end{center}
\end{table*}

\subsection{Table-top Object Category Recognition System}
We built a near real-time system for category recognition of table-top objects based on the homeomorphic manifold analysis framework described. Our system was trained on a subset of 10 different categories from the RGB-D Object dataset. The category recognition runtime per object in one frame is less than 2 seconds. Our MATLAB implementation was not optimized for real-time processing but despite this, the potential for real-time capability is evident. We only performed visual-only and depth-only training and testing of the system. We did not experiment with the combination of both modes as we wanted to optimize for speed as much as possible.

\begin{figure}[h]
\centering
\includegraphics[width=1.0\linewidth]{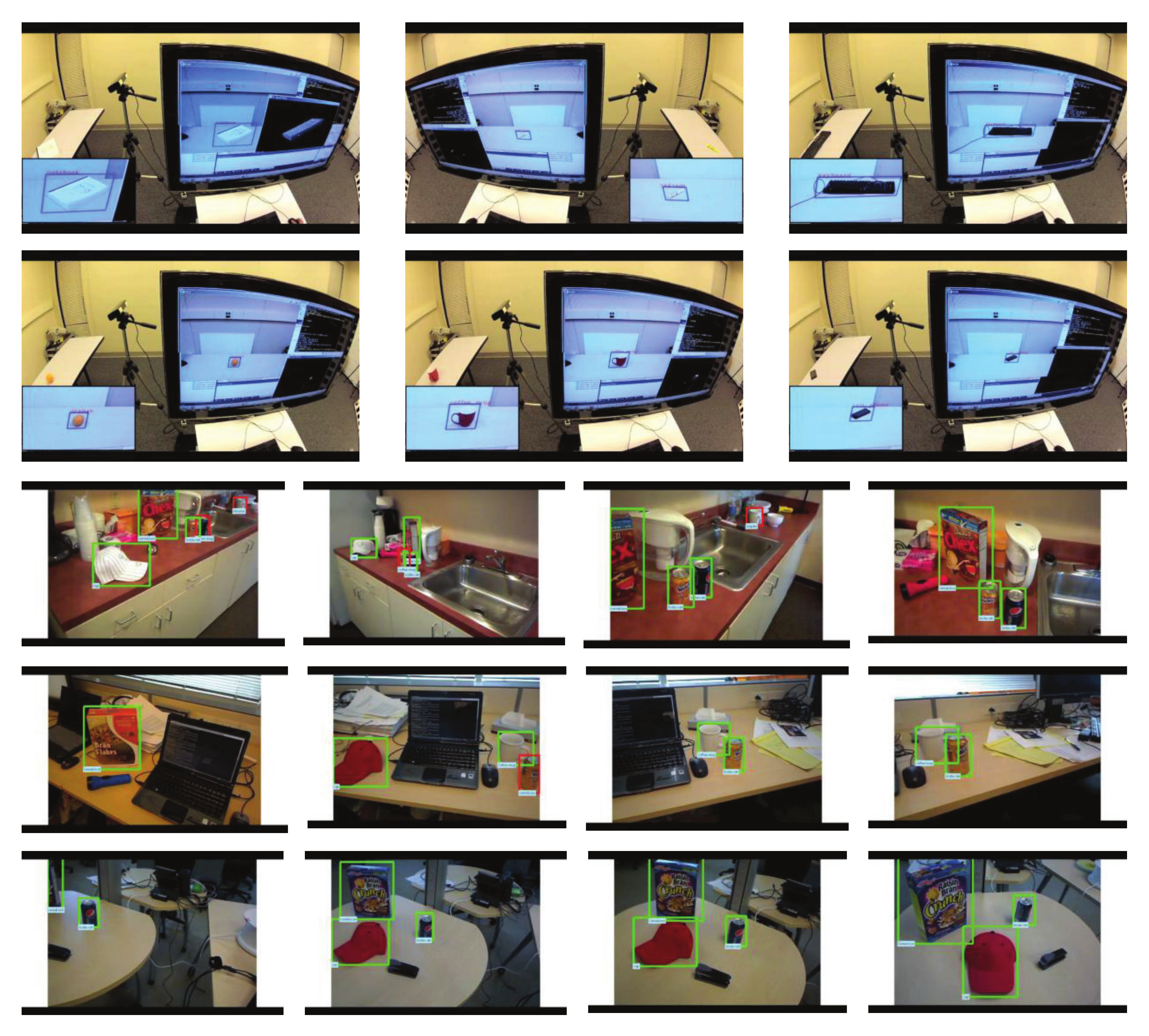}
\caption{Near real-time system running on single table-top objects (first 2 rows) and the RGB-D Scenes dataset (last 3 rows, where green boxes indicate correct results while red boxes indicate incorrect results). }
\label{fig:system}
\end{figure}

The system was tested on videos provided in the RGB-D Scenes dataset that contain cluttered scenes with occlusion, a much larger variation of viewpoint and varying scales (\emph{e.g.} kitchen and desk scenes). Our system achieved \textgreater$62\%$ category recognition accuracy using the depth mode only. An interesting observation was that depth-only recognition outperformed visual-only recognition in these cluttered scenes; intuitively due to the fact that background texture around objects introduces visual noise. In the depth mode, large depth discontinuities help to separate objects from background clutter and this aids recognition. We also tested our system on never-seen-before objects placed on table-tops without clutter. For this, we used the visual-only mode since there was no clutter in the scene. Fig.~\ref{fig:system} shows results achieved by our system running on never-seen-before objects and objects from the videos provided in the RGB-D Scenes dataset.

Depth segmentation was performed on point clouds generated using the Kinect sensor in real-time using the Point Cloud Library \cite{Rusu2011pcl}. This allows the table-top object to be segmented away from the table plane. The segmented objects are then found in the visual and depth images using the segmented object in the point cloud and then cropped to the size of the object. We then perform category recognition on these cropped images.

\subsection{Computational Complexity}
The computation complexity of SVD scales cubically with the number of objects in our case ($O(N^3)$). SVD can be done offline on large enough matrices containing tens of thousands of rows. The running time of our near real-time system for table-top object category recognition also shows that the computational complexity of the estimation phase is acceptable and has the potential for real-time.

\section{Conclusion}
\label{Sec:concl}
In this work we have presented a unified framework that is based on homeomorphic mapping between a common manifold and object manifolds in order to jointly solve the 3 subproblems of object recognition: category, instance and pose recognition. Extensive experiments on several recent and large datasets validates the robustness and strength of our approach. We significantly outperform state-of-the-art in pose recognition. For object recognition we achieve accuracy on par and in some cases better than state-of-the-art. We have also shown the capability of our approach in estimating full 3D pose. We have also shown the potential for real-time application to AI and robotic visual reasoning by building a working near real-time system that performs table-top object detection and category recognition using the Kinect sensor.



\section*{Acknowledgment}
This work was partly supported by the Office of Navel Research grant N00014-12-1-0755.
This work was also partly supported by the National Natural Science Foundation of China (No. 61371134 and No.
61071137), the National Basic Research Program of China (Project No. 2010CB327900), and the Fundamental Research Funds for the Central Universities (YWF-14-RSC-115).

\section*{References}

\bibliography{allbibfile}

\begin{thebibliography}{10}
\expandafter\ifx\csname url\endcsname\relax
  \def\url#1{\texttt{#1}}\fi
\expandafter\ifx\csname urlprefix\endcsname\relax\def\urlprefix{URL }\fi
\expandafter\ifx\csname href\endcsname\relax
  \def\href#1#2{#2} \def\path#1{#1}\fi

\bibitem{ZhangAAAI13}
H.~Zhang, T.~El-Gaaly, A.~Elgammal, Z.~Jiang, Joint object and pose recognition
  using {H}omeomorphic {M}anifold {A}nalysis, in: Twenty-Seventh AAAI
  Conference on Artificial Intelligence, 2013, pp. 1012--1019.

\bibitem{Marr82}
D.~Marr, Vision: A computational investigation into the human representation
  and processing of visual information, W.H. Freeman and Company, 1982.

\bibitem{grimson1985recognition}
W.~Grimson, T.~Lozano-Perez, Recognition and localization of overlapping parts
  from sparse data in two and three dimensions, in: Robotics and Automation.
  Proceedings. 1985 IEEE International Conference on, Vol.~2, IEEE, 1985, pp.
  61--66.
\newblock \href {http://dx.doi.org/10.1109/ROBOT.1985.1087320}
  {\path{doi:10.1109/ROBOT.1985.1087320}}.

\bibitem{lamdan1988geometric}
Y.~Lamdan, H.~Wolfson, Geometric hashing: A general and efficient model-based
  recognition scheme, in: Computer Vision., Second International Conference on,
  1988, pp. 238--249.
\newblock \href {http://dx.doi.org/10.1109/CCV.1988.589995}
  {\path{doi:10.1109/CCV.1988.589995}}.

\bibitem{lowe1987three}
D.~G. Lowe, Three-dimensional object recognition from single two-dimensional
  images, Artificial intelligence 31~(3) (1987) 355--395.
\newblock \href {http://dx.doi.org/10.1016/0004-3702(87)90070-1}
  {\path{doi:10.1016/0004-3702(87)90070-1}}.

\bibitem{shimshoni1997finite}
I.~Shimshoni, J.~Ponce, Finite-resolution aspect graphs of polyhedral objects,
  Pattern Analysis and Machine Intelligence, IEEE Transactions on 19~(4) (1997)
  315--327.
\newblock \href {http://dx.doi.org/10.1109/34.588001}
  {\path{doi:10.1109/34.588001}}.

\bibitem{Felzen05}
P.~F. Felzenszwalb, D.~P. Huttenlocher, Pictorial structures for object
  recognition, International Journal of Computer Vision 61~(1) (2005) 55--79.
\newblock \href {http://dx.doi.org/10.1023/B:VISI.0000042934.15159.49}
  {\path{doi:10.1023/B:VISI.0000042934.15159.49}}.

\bibitem{pedro2010}
P.~F. Felzenszwalb, R.~B. Girshick, D.~McAllester, D.~Ramanan, Object detection
  with discriminatively trained part-based models, Pattern Analysis and Machine
  Intelligence, IEEE Transactions on 32~(9) (2010) 1627--1645.
\newblock \href {http://dx.doi.org/10.1109/TPAMI.2009.167}
  {\path{doi:10.1109/TPAMI.2009.167}}.

\bibitem{bof_nineclasses04}
J.~Willamowski, D.~Arregui, G.~Csurka, C.~R. Dance, L.~Fan, Categorizing nine
  visual classes using local appearance descriptors, in: ICPR 2004 Workshop
  Learning for Adaptable Visual Systems Cambridge, 2004.

\bibitem{bof_iccv05}
J.~Sivic, B.~Russell, A.~Efros, A.~Zisserman, W.~Freeman, Discovering objects
  and their location in images, in: IEEE International Conference on Computer
  Vision, Vol.~1, 2005, pp. 370--377.
\newblock \href {http://dx.doi.org/10.1109/ICCV.2005.77}
  {\path{doi:10.1109/ICCV.2005.77}}.

\bibitem{Chiu07}
H.-P. Chiu, L.~Kaelbling, T.~Lozano-Perez, Virtual training for multi-view
  object class recognition, in: IEEE Conference on Computer Vision and Pattern
  Recognition, 2007, pp. 1--8.
\newblock \href {http://dx.doi.org/10.1109/CVPR.2007.383044}
  {\path{doi:10.1109/CVPR.2007.383044}}.

\bibitem{Kushal07}
A.~Kushal, C.~Schmid, J.~Ponce, Flexible object models for category-level 3d
  object recognition, in: IEEE Conference on Computer Vision and Pattern
  Recognition, 2007, pp. 1--8.
\newblock \href {http://dx.doi.org/10.1109/CVPR.2007.383149}
  {\path{doi:10.1109/CVPR.2007.383149}}.

\bibitem{Savarese07}
S.~Savarese, L.~Fei-Fei, 3d generic object categorization, localization and
  pose estimation, in: IEEE 11th International Conference on Computer Vision,
  2007, pp. 1--8.
\newblock \href {http://dx.doi.org/10.1109/ICCV.2007.4408987}
  {\path{doi:10.1109/ICCV.2007.4408987}}.

\bibitem{Liebelt08}
J.~Liebelt, C.~Schmid, K.~Schertler, Viewpoint-independent object class
  detection using 3d feature maps, in: IEEE Conference on Computer Vision and
  Pattern Recognition, 2008, pp. 1--8.
\newblock \href {http://dx.doi.org/10.1109/CVPR.2008.4587614}
  {\path{doi:10.1109/CVPR.2008.4587614}}.

\bibitem{Silvio09}
H.~Su, M.~Sun, L.~Fei-Fei, S.~Savarese, Learning a dense multi-view
  representation for detection, viewpoint classification and synthesis of
  object categories, in: IEEE 12th International Conference on Computer Vision,
  2009, pp. 213--220.
\newblock \href {http://dx.doi.org/10.1109/ICCV.2009.5459168}
  {\path{doi:10.1109/ICCV.2009.5459168}}.

\bibitem{sun09}
M.~Sun, H.~Su, S.~Savarese, L.~Fei~Fei, A multi-view probabilistic model for
  3{D} object classes, in: IEEE Conference on Computer Vision and Pattern
  Recognition, 2009, pp. 1247--1254.
\newblock \href {http://dx.doi.org/10.1109/CVPR.2009.5206723}
  {\path{doi:10.1109/CVPR.2009.5206723}}.

\bibitem{Payet2011}
N.~Payet, S.~Todorovic, From contours to 3{D} object detection and pose
  estimation, in: IEEE International Conference on Computer Vision, 2011, pp.
  983--990.
\newblock \href {http://dx.doi.org/10.1109/ICCV.2011.6126342}
  {\path{doi:10.1109/ICCV.2011.6126342}}.

\bibitem{mei2011robust}
L.~Mei, J.~Liu, A.~Hero, S.~Savarese, Robust object pose estimation via
  statistical manifold modeling, in: IEEE International Conference on Computer
  Vision, 2011, pp. 967--974.
\newblock \href {http://dx.doi.org/10.1109/ICCV.2011.6126340}
  {\path{doi:10.1109/ICCV.2011.6126340}}.

\bibitem{HMA}
A.~Elgammal, C.-S. Lee, Homeomorphic manifold analysis ({HMA}): Generalized
  separation of style and content on manifolds, Image and Vision Computing
  31~(4) (2013) 291--310.
\newblock \href {http://dx.doi.org/10.1016/j.imavis.2012.12.003}
  {\path{doi:10.1016/j.imavis.2012.12.003}}.

\bibitem{tamjidi2013}
A.~Tamjidi, C.~Ye, S.~Hong, 6-{DOF} pose estimation of a portable navigation
  aid for the visually impaired, in: IEEE International Symposium on Robotic
  and Sensors Environments, 2013, pp. 178--183.
\newblock \href {http://dx.doi.org/10.1109/ROSE.2013.6698439}
  {\path{doi:10.1109/ROSE.2013.6698439}}.

\bibitem{Savarese08}
S.~Savarese, L.~Fei-Fei, View synthesis for recognizing unseen poses of object
  classes, in: D.~Forsyth, P.~Torr, A.~Zisserman (Eds.), Computer Vision –
  ECCV 2008, Vol. 5304 of Lecture Notes in Computer Science, Springer Berlin
  Heidelberg, 2008, pp. 602--615.
\newblock \href {http://dx.doi.org/10.1007/978-3-540-88690-7_45}
  {\path{doi:10.1007/978-3-540-88690-7_45}}.

\bibitem{schels2012learning}
J.~Schels, J.~Liebelt, R.~Lienhart, Learning an object class representation on
  a continuous viewsphere, in: Computer Vision and Pattern Recognition (CVPR),
  2012 IEEE Conference on, 2012, pp. 3170--3177.
\newblock \href {http://dx.doi.org/10.1109/CVPR.2012.6248051}
  {\path{doi:10.1109/CVPR.2012.6248051}}.

\bibitem{Pepik12}
B.~Pepik, M.~Stark, P.~Gehler, B.~Schiele, Teaching 3d geometry to deformable
  part models, in: IEEE Conference on Computer Vision and Pattern Recognition,
  IEEE, 2012, pp. 3362--3369.
\newblock \href {http://dx.doi.org/10.1109/CVPR.2012.6248075}
  {\path{doi:10.1109/CVPR.2012.6248075}}.

\bibitem{Guo2008}
G.~Guo, Y.~Fu, C.~Dyer, T.~Huang, Head pose estimation: Classification or
  regression?, in: International Conference on Pattern Recognition, 2008, pp.
  1--4.
\newblock \href {http://dx.doi.org/10.1109/ICPR.2008.4761081}
  {\path{doi:10.1109/ICPR.2008.4761081}}.

\bibitem{Ando2005}
S.~Ando, Y.~Kusachi, A.~Suzuki, K.~Arakawa, Appearance based pose estimation of
  3{D} object using support vector regression, in: IEEE International
  Conference on Image Processing, 2005, pp. I--341--4.
\newblock \href {http://dx.doi.org/10.1109/ICIP.2005.1529757}
  {\path{doi:10.1109/ICIP.2005.1529757}}.

\bibitem{Torki11}
M.~Torki, A.~Elgammal, Regression from local features for viewpoint and pose
  estimation, in: IEEE International Conference on Computer Vision, 2011, pp.
  2603--2610.
\newblock \href {http://dx.doi.org/10.1109/ICCV.2011.6126549}
  {\path{doi:10.1109/ICCV.2011.6126549}}.

\bibitem{Fanelli2011headpose}
G.~Fanelli, T.~Weise, J.~Gall, L.~Van~Gool, Real time head pose estimation from
  consumer depth cameras, in: R.~Mester, M.~Felsberg (Eds.), Pattern
  Recognition, Vol. 6835 of Lecture Notes in Computer Science, Springer Berlin
  Heidelberg, 2011, pp. 101--110.
\newblock \href {http://dx.doi.org/10.1007/978-3-642-23123-0_11}
  {\path{doi:10.1007/978-3-642-23123-0_11}}.

\bibitem{ElGaaly2012rgbd_mkl}
T.~El-Gaaly, M.~Torki, A.~Elgammal, M.~Singh, {RGBD} object pose recognition
  using local-global multi-kernel regression, in: International Conference on
  Pattern Recognition, 2012, pp. 2468--2471.

\bibitem{Pan2013headpose}
L.~Pan, R.~Liu, M.~Xie, Mixture of related regressions for head pose
  estimation, in: IEEE International Conference on Image Processing, 2013, pp.
  3647--3651.
\newblock \href {http://dx.doi.org/10.1109/ICIP.2013.6738752}
  {\path{doi:10.1109/ICIP.2013.6738752}}.

\bibitem{Lai2011optree}
K.~Lai, L.~Bo, X.~Ren, D.~Fox, A scalable tree-based approach for joint object
  and pose recognition, in: Twenty-Fifth AAAI Conference on Artificial
  Intelligenc, 2011, pp. 1474--1480.

\bibitem{Fanelli2013randomforests}
G.~Fanelli, M.~Dantone, J.~Gall, A.~Fossati, L.~Van~Gool, Random forests for
  real time 3d face analysis, International Journal of Computer Vision 101~(3)
  (2013) 437--458.
\newblock \href {http://dx.doi.org/10.1007/s11263-012-0549-0}
  {\path{doi:10.1007/s11263-012-0549-0}}.

\bibitem{murase95visual}
H.~Murase, S.~Nayar, Visual learning and recognition of 3d objects from
  appearance, International Journal of Computer Vision 14 (1995) 5--24.
\newblock \href {http://dx.doi.org/10.1007/BF01421486}
  {\path{doi:10.1007/BF01421486}}.

\bibitem{Elgammal04CVPRb}
A.~Elgammal, C.-S. Lee, Inferring 3d body pose from silhouettes using activity
  manifold learning, in: IEEE Conference on Computer Vision and Pattern
  Recognition, Vol.~2, 2004, pp. 681--688.
\newblock \href {http://dx.doi.org/10.1109/CVPR.2004.1315230}
  {\path{doi:10.1109/CVPR.2004.1315230}}.

\bibitem{Urtasun06CVPR}
R.~Urtasun, D.~Fleet, P.~Fua, 3{D} people tracking with gaussian process
  dynamical models, in: IEEE Computer Society Conference on Computer Vision and
  Pattern Recognition, Vol.~1, 2006, pp. 238--245.
\newblock \href {http://dx.doi.org/10.1109/CVPR.2006.15}
  {\path{doi:10.1109/CVPR.2006.15}}.

\bibitem{Jolliffe86PCA}
I.~T. Jolliffe, Principal Component Analysis, Springer-Verlag, 1986.

\bibitem{Tenenbaum00}
J.~B. Tenenbaum, W.~T. Freeman, Separating style and content with bilinear
  models, Neural Computation 12~(6) (2000) 1247--1283.
\newblock \href {http://dx.doi.org/10.1162/089976600300015349}
  {\path{doi:10.1162/089976600300015349}}.

\bibitem{Vasilescu02}
M.~Vasilescu, D.~Terzopoulos, Multilinear analysis of image ensembles:
  Tensorfaces, in: A.~Heyden, G.~Sparr, M.~Nielsen, P.~Johansen (Eds.),
  Computer Vision — ECCV 2002, Vol. 2350 of Lecture Notes in Computer
  Science, Springer Berlin Heidelberg, 2002, pp. 447--460.
\newblock \href {http://dx.doi.org/10.1007/3-540-47969-4_30}
  {\path{doi:10.1007/3-540-47969-4_30}}.

\bibitem{Lathauwer00JMAAa}
L.~De~Lathauwer, B.~De~Moor, J.~Vandewalle, A multilinear singular value
  decomposition, SIAM Journal on Matrix Analysis and Applications 21~(4) (2000)
  1253--1278.
\newblock \href {http://dx.doi.org/10.1137/S0895479896305696}
  {\path{doi:10.1137/S0895479896305696}}.

\bibitem{Kapteyn86PSY}
A.~Kapteyn, H.~Neudecker, T.~Wansbeek, An approach to n-model component
  analysis, Psychometrika 51~(2) (1986) 269--275.
\newblock \href {http://dx.doi.org/10.1007/BF02293984}
  {\path{doi:10.1007/BF02293984}}.

\bibitem{Magnus88}
J.~R. Magnus, H.~Neudecker, Matrix Differential Calculus with Applications in
  Statistics and Econometrics, John Wiley \& Sons, 1988.

\bibitem{Roweis00}
S.~Roweis, L.~Saul, Nonlinear dimensionality reduction by locally linear
  embedding, Sciene 290~(5500) (2000) 2323--2326.
\newblock \href {http://dx.doi.org/10.1126/science.290.5500.2323}
  {\path{doi:10.1126/science.290.5500.2323}}.

\bibitem{Tenenbaum98}
J.~B. Tenenbaum, Mapping a manifold of perceptual observations, in: M.~Jordan,
  M.~Kearns, S.~Solla (Eds.), Advances in Neural Information Processing Systems
  10, MIT Press, 1998, pp. 682--688.

\bibitem{Belkin03NC}
M.~Belkin, P.~Niyogi, Laplacian eigenmaps for dimensionality reduction and data
  representation, Neural Computation 15~(6) (2003) 1373--1396.
\newblock \href {http://dx.doi.org/10.1162/089976603321780317}
  {\path{doi:10.1162/089976603321780317}}.

\bibitem{Brand03}
M.~Brand, K.~Huang, A unifying theorem for spectral embedding and clustering,
  in: Proc. of the Ninth International Workshop on AI and Statistics, 2003.

\bibitem{lawrence03NIPS}
N.~D. Lawrence, Gaussian process latent variable models for visualisation of
  high dimensional data, in: S.~Thrun, L.~Saul, B.~Sch\"{o}lkopf (Eds.),
  Advances in Neural Information Processing Systems 16, MIT Press, 2004, pp.
  329--336.

\bibitem{Weinberger04CVPR}
K.~Weinberger, L.~Saul, Unsupervised learning of image manifolds by
  semidefinite programming, in: IEEE Computer Society Conference on Computer
  Vision and Pattern Recognition, Vol.~2, 2004, pp. 988--995.
\newblock \href {http://dx.doi.org/10.1109/CVPR.2004.1315272}
  {\path{doi:10.1109/CVPR.2004.1315272}}.

\bibitem{Kimeldorf1970:representerThrm}
G.~S. Kimeldorf, G.~Wahba, A correspondence between bayesian estimation on
  stochastic processes and smoothing by splines, The Annals of Mathematical
  Statistics 41~(2) (1970) 495--502.
\newblock \href {http://dx.doi.org/10.1214/aoms/1177697089}
  {\path{doi:10.1214/aoms/1177697089}}.

\bibitem{Poggio1990:GRBF}
T.~Poggio, F.~Girosi, Networks for approximation and learning, Proceedings of
  the IEEE 78~(9) (1990) 1481--1497.
\newblock \href {http://dx.doi.org/10.1109/5.58326}
  {\path{doi:10.1109/5.58326}}.

\bibitem{Tipping99PPCA}
M.~E. Tipping, C.~M. Bishop, Probabilistic principal component analysis,
  Journal of the Royal Statistical Society: Series B (Statistical Methodology)
  61~(3) (1999) 611--622.
\newblock \href {http://dx.doi.org/10.1111/1467-9868.00196}
  {\path{doi:10.1111/1467-9868.00196}}.

\bibitem{Ham05}
J.~H. Ham, D.~D. Lee, L.~K. Saul, Semisupervised alignment of manifolds, in:
  Proceedings of the Tenth International Workshop on Artificial Intelligence
  and Statistics, 2005, pp. 120--127.

\bibitem{Wahba71splines}
G.~Kimeldorf, G.~Wahba, Some results on tchebycheffian spline functions,
  Journal of Mathematical Analysis and Applications 33~(1) (1971) 82--95.
\newblock \href {http://dx.doi.org/10.1016/0022-247X(71)90184-3}
  {\path{doi:10.1016/0022-247X(71)90184-3}}.

\bibitem{elgammal04cvpr}
A.~Elgammal, C.-S. Lee, Separating style and content on a nonlinear manifold,
  in: IEEE Conference on Computer Vision and Pattern Recognition, Vol.~1, 2004,
  pp. 478--485.
\newblock \href {http://dx.doi.org/10.1109/CVPR.2004.1315070}
  {\path{doi:10.1109/CVPR.2004.1315070}}.

\bibitem{Arulampalam02pf}
M.~Arulampalam, S.~Maskell, N.~Gordon, T.~Clapp, A tutorial on particle filters
  for online nonlinear/non-gaussian bayesian tracking, IEEE Transactions on
  Signal Processing 50~(2) (2002) 174--188.
\newblock \href {http://dx.doi.org/10.1109/78.978374}
  {\path{doi:10.1109/78.978374}}.

\bibitem{COIL20}
S.~A. Nene, S.~K. Nayar, H.~Murase, Columbia object image library ({COIL}-20),
  Tech. Rep. CUCS-005-96, Columbia University (1996).

\bibitem{Ozuysal2009pose}
M.~Ozuysal, V.~Lepetit, P.~Fua, Pose estimation for category specific multiview
  object localization, in: IEEE Conference on Computer Vision and Pattern
  Recognition 2009, 2009, pp. 778--785.
\newblock \href {http://dx.doi.org/10.1109/CVPR.2009.5206633}
  {\path{doi:10.1109/CVPR.2009.5206633}}.

\bibitem{sun2010depth}
M.~Sun, G.~Bradski, B.-X. Xu, S.~Savarese, Depth-encoded hough voting for joint
  object detection and shape recovery, in: K.~Daniilidis, P.~Maragos,
  N.~Paragios (Eds.), Computer Vision – ECCV 2010, Vol. 6315 of Lecture Notes
  in Computer Science, Springer Berlin Heidelberg, 2010, pp. 658--671.
\newblock \href {http://dx.doi.org/10.1007/978-3-642-15555-0_48}
  {\path{doi:10.1007/978-3-642-15555-0_48}}.

\bibitem{Yu2014wacv}
Y.~Xiang, R.~Mottaghi, S.~Savarese, Beyond pascal: A benchmark for 3d object
  detection in the wild, in: IEEE Winter Conference on Applications of Computer
  Vision, 2014, pp. 75--82.
\newblock \href {http://dx.doi.org/10.1109/WACV.2014.6836101}
  {\path{doi:10.1109/WACV.2014.6836101}}.

\bibitem{Lai2011rgbddataset}
K.~Lai, L.~Bo, X.~Ren, D.~Fox, A large-scale hierarchical multi-view rgb-d
  object dataset, in: 2011 IEEE International Conference on Robotics and
  Automation (ICRA), 2011, pp. 1817 --1824.
\newblock \href {http://dx.doi.org/10.1109/ICRA.2011.5980382}
  {\path{doi:10.1109/ICRA.2011.5980382}}.

\bibitem{Dalal05hog}
N.~Dalal, B.~Triggs, Histograms of oriented gradients for human detection, in:
  IEEE Conference on Computer Vision and Pattern Recognition 2005, Vol.~1,
  2005, pp. 886--893.
\newblock \href {http://dx.doi.org/10.1109/CVPR.2005.177}
  {\path{doi:10.1109/CVPR.2005.177}}.

\bibitem{SavareseFeiFei2009}
M.~Sun, H.~Su, S.~Savarese, L.~Fei-Fei, A multi-view probabilistic model for 3d
  object classes, in: IEEE Conference on Computer Vision and Pattern
  Recognition 2009, 2009, pp. 1247--1254.
\newblock \href {http://dx.doi.org/10.1109/CVPR.2009.5206723}
  {\path{doi:10.1109/CVPR.2009.5206723}}.

\bibitem{schmid20103Dgeometricmodels}
J.~Liebelt, C.~Schmid, Multi-view object class detection with a 3d geometric
  model, in: IEEE Conference on Computer Vision and Pattern Recognition 2010,
  2010, pp. 1688--1695.
\newblock \href {http://dx.doi.org/10.1109/CVPR.2010.5539836}
  {\path{doi:10.1109/CVPR.2010.5539836}}.

\bibitem{bakry2014untangling}
A.~Bakry, A.~Elgammal, Untangling object-view manifold for multiview
  recognition and pose estimation, in: Computer Vision--ECCV 2014, Springer,
  2014, pp. 434--449.

\bibitem{Rusu2010VFH}
R.~Rusu, G.~Bradski, R.~Thibaux, J.~Hsu, Fast 3d recognition and pose using the
  viewpoint feature histogram, in: 2010 IEEE/RSJ International Conference on
  Intelligent Robots and Systems (IROS), 2010, pp. 2155--2162.
\newblock \href {http://dx.doi.org/10.1109/IROS.2010.5651280}
  {\path{doi:10.1109/IROS.2010.5651280}}.

\bibitem{Rusu2011pcl}
R.~B. Rusu, S.~Cousins, {3D is here: Point Cloud Library (PCL)}, in: IEEE
  International Conference on Robotics and Automation, ICRA 2011, Shanghai,
  China, 9-13 May 2011, IEEE, 2011.
\newblock \href {http://dx.doi.org/10.1109/ICRA.2011.5980567}
  {\path{doi:10.1109/ICRA.2011.5980567}}.

\end{thebibliography}

\end{document}